\documentclass[10pt,twocolumn,letterpaper]{article}
\usepackage[pagenumbers]{cvpr}

\usepackage{graphicx}
\usepackage{amsmath}
\usepackage{amssymb}
\usepackage{booktabs}
\usepackage{url}
\usepackage{multirow}
\usepackage{array}
\usepackage{color}
\usepackage{bbding}
\usepackage{xspace}
\usepackage{wrapfig}
\usepackage{pifont}
\usepackage{colortbl}
\usepackage{algorithm}
\usepackage{algpseudocode}
\usepackage{indentfirst} 
\usepackage{makecell}
\usepackage{cancel}
\usepackage{listings}
\usepackage{etoolbox}
\usepackage[pagebackref,breaklinks,colorlinks]{hyperref}

\usepackage[capitalize]{cleveref}
\crefname{section}{Sec.}{Secs.}
\Crefname{section}{Section}{Sections}
\Crefname{table}{Table}{Tables}
\crefname{table}{Tab.}{Tabs.}

\makeatletter
\AfterEndEnvironment{algorithm}{\let\@algcomment\relax}
\AtEndEnvironment{algorithm}{\kern2pt\hrule\relax\vskip3pt\@algcomment}
\let\@algcomment\relax
\newcommand\algcomment[1]{\def\@algcomment{\footnotesize#1}}
\renewcommand\fs@ruled{\def\@fs@cfont{\bfseries}\let\@fs@capt\floatc@ruled
  \def\@fs@pre{\hrule height.8pt depth0pt \kern2pt}%
  \def\@fs@post{}%
  \def\@fs@mid{\kern2pt\hrule\kern2pt}%
  \let\@fs@iftopcapt\iftrue}
\makeatother

\def\cvprPaperID{****}
\def\confName{****}
\def\confYear{****}

\definecolor{linkcolor}{RGB}{218, 50, 138}
\begin{document}

\title{Self-Supervised Learning by Estimating Twin Class Distributions}
\author{
        Feng Wang$ ^1$\thanks{This work was done during an internship at ByteDance. Correspondence to: Tao Kong and Huaping Liu}
        ~~~~~
        Tao Kong$ ^{2} $ 
        ~~~~~
        Rufeng Zhang$ ^{3*}$
        ~~~~~
        Huaping Liu$ ^1  $
        ~~~~~
        Hang Li$ ^{2}$ 
        \\~
        $ ^1 $Tsinghua University
        ~~~~
        $ ^2 $ByteDance AI Lab  
        ~~~~
        $ ^3 $Tongji University 
        \\
        \texttt{\{wang-f20@mails., hpliu@\}tsinghua.edu.cn}\\
        \texttt{\{kongtao, lihang.lh\}@bytedance.com}\\
        \texttt{cxrfzhang@tongji.edu.cn}
}

\maketitle
\def\ourmethod{\textsc{Twist}\xspace}
\begin{abstract}
We present \ourmethod, a simple and theoretically explainable self-supervised representation learning method by classifying large-scale unlabeled datasets in an end-to-end way. We employ a siamese network terminated by a softmax operation to produce twin class distributions of two augmented images. Without supervision, we enforce the class distributions of different augmentations to be consistent. However, simply minimizing the divergence between augmentations will cause collapsed solutions, i.e., outputting the same class probability distribution for all images. In this case, no information about the input image 
is left. To solve this problem, we propose to maximize the mutual information between the input and the class predictions. Specifically, we minimize the entropy of the distribution for each sample to make the class prediction for each sample assertive and maximize the entropy of the mean distribution to make the predictions of different samples diverse. In this way, \ourmethod can naturally avoid the collapsed solutions without specific designs such as asymmetric network, stop-gradient operation, or momentum encoder. 
As a result, \ourmethod outperforms state-of-the-art methods on a wide range of tasks. Especially, \ourmethod performs surprisingly well on semi-supervised learning, achieving $61.2\%$ 
top-1 accuracy with $1\%$ 
ImageNet labels using a ResNet-50 as backbone, surpassing previous best results by an absolute improvement of $6.2\%$. Codes and pre-trained models are given on: \textcolor{linkcolor}{https://github.com/bytedance/TWIST}

\end{abstract}

\section{Introduction}
\label{sec:intro}
Deep neural networks learned from large-scale datasets have powered many aspects of machine learning.
In computer vision, the neural networks trained on the ImageNet dataset~\cite{deng2009imagenet} can perform better than or as well as humans in image classification~\cite{krizhevsky2012imagenet,szegedy2015going,he2016deep,dosovitskiy2020image}. 
The obtained representations can also be adapted to other downstream tasks
\cite{ren2015faster,he2017mask,long2015fully}.
However, learning from large-scale labeled data requires expensive annotations, making it difficult to scale.
\begin{figure*}[t]
  \begin{center}
  \includegraphics[width=1.0\linewidth]{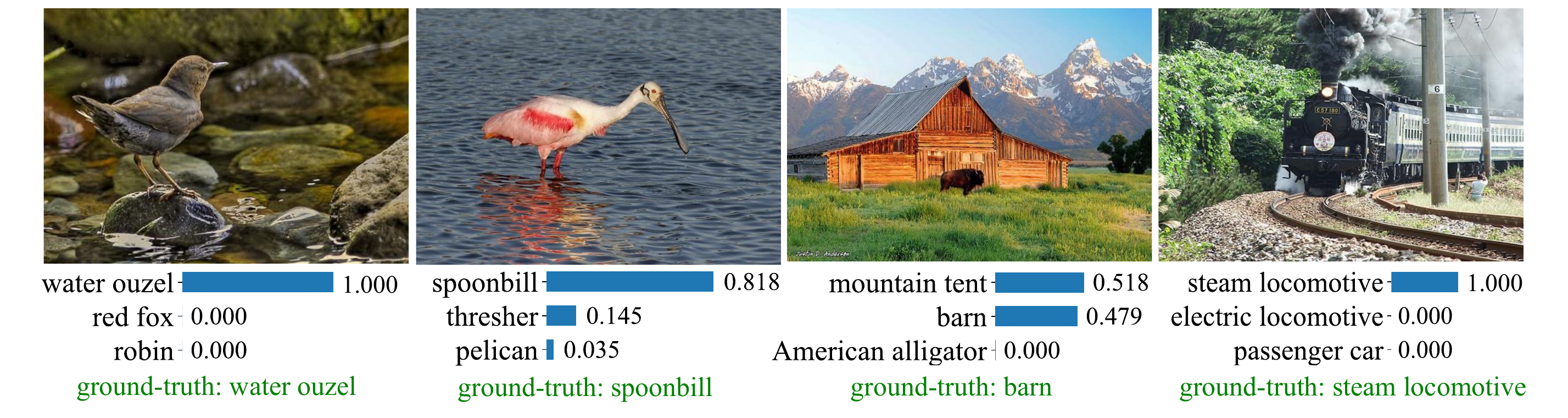}
  \end{center}
  \vspace{-0.6cm}
  \caption{Examples of unsupervised top-3 predictions of \ourmethod. The predicted class indices are mapped to the labels in ImageNet by Kuhn–Munkres algorithm \cite{kuhn1955hungarian}. Note that the labels are only used to map our predictions to the ImageNet labels, we do not use any label to participate in the training process. More examples are given in Appendix.}
  \label{head}
  \vspace{-0.3cm}
\end{figure*}

Recently, self-supervised learning has achieved remarkable progress and largely closed the gap with supervised learning. The contrastive learning approaches ~\cite{wu2018unsupervised, he2020momentum, chen2020simple, chen2020improved, grill2020bootstrap, chen2021exploring} learn representations by maximizing the agreement of different augmentations and pushing away the representations of different images based on the instance discrimination pretext task \cite{wu2018unsupervised}. BYOL \cite{grill2020bootstrap} and SimSiam \cite{chen2021exploring} propose to abandon the negative samples, and design the asymmetric architecture and momentum encoder (or stop-gradient) to avoid collapsed solutions.
Clustering-based methods
\cite{caron2018deep, asano2019self, caron2020unsupervised} 
usually employ the clustering algorithms to generate supervision signals 
to guide the learning process, and constitute an unsupervised classification task. 

This work focuses on the unsupervised classification pretext task and explores learning representations by classifying unlabeled images end-to-end. Without labels as accurate supervision, we learn from recent successful self-supervised methods \cite{he2020momentum, chen2020simple, caron2020unsupervised, caron2021emerging} that adequately utilize the consistency between augmentations. In the unsupervised classification pretext tasks \cite{caron2020unsupervised, caron2021emerging}, this is reflected by requiring different augmentations to have identical predictions.
However, simply minimizing the divergence between the predictions of different augmentations will cause the collapsed solution problem, i.e., outputting the same class for all images. To solve this problem, clustering-based methods \cite{caron2018deep, asano2019self, li2020prototypical, caron2019unsupervised, caron2020unsupervised} adopt the clustering techniques such as K-means \cite{caron2018deep, li2020prototypical}, and Sinkhorn-Knopp algorithm \cite{asano2019self, caron2020unsupervised} to generate 
target assignments for each image as supervision. DINO \cite{caron2021emerging} proposes sharpening and centering technique, together with momentum encoder, to avoid generating collapsed solutions. 
Recent Self-Classifier \cite{amrani2021self} 
avoids collapsed solutions by asserting the uniform prior to the predicted classes. 

In this paper, we propose to solve the collapsed solution problem by maximizing the mutual information between input images and output predictions, and our method is named \ourmethod (Twin Class Distribution Estimation). The motivation is based on the observation that the collapsed solution (i.e., outputting the same class for all images) will cause the class predictions to carry no information about the input images. Thus an intuitive and straightforward way is to maximize the shared information between image $X$ and class prediction $Y$. The mutual information maximization between input and class predictions is a historical method which can be traced back to 30 years ago --- \textit{Unsupervised  classifiers,  mutual  information  and `phantom targets'} \cite{bridle1992unsupervised}. To the best of our knowledge, we are the first to successfully apply it to large-scale representation learning. Specifically, we minimize the entropy of the class distribution for each sample to make the class distribution sharp and maximize the entropy of the \textit{mean class distribution} to make the predictions for different samples diverse. We theoretically prove that maximizing the mutual information is equal to the combination of the above two intuitive objectives. The above optimizing problem can be achieved by a unified objective function -- \ourmethod loss. This makes \ourmethod more straightforward, eliminating the reliance on clustering techniques or complicated architecture designs. Besides, compared with clustering-based methods and DINO which explicitly generate targets for images, the target of \ourmethod is like a `\textit{phantom}' that is not known before optimizing and is decided by the cooperative learning of different views.

\ourmethod can not only successfully classify unlabeled images (Fig. \ref{head}), but also derives high-quality features. We evaluate the performance of \ourmethod on many downstream tasks, surpassing state-of-the-art methods on a wide range of tasks including semi-supervised learning (+6.2 top-1 accuracy on 1\% labeled data compared with previous best method), linear classification, transfer learning and dense tasks. These results show that \ourmethod successfully connects unsupervised classification and representation learning, and could serve as a strong baseline for both purposes. Overall, the contributions are summarized as follows:
\begin{itemize}
    \item 
    We propose a straightforward, theoretical explainable self-supervised learning method by classifying large-scale unlabeled datasets, eliminating the requirements of complicated architecture designs.
    \item 
    We show that \ourmethod itself is an efficient unsupervised classifier. Without special adaptation, we achieve the best unsupervised classification results compared with other methods.
    \item 
    The representation quality of \ourmethod has been evaluated on various downstream tasks, including ImageNet linear and fine-tuning settings, semi-supervised learning, and dense predictive tasks, achieving state-of-the-art results on most of them. The results indicate that \ourmethod can serve as a general pre-training method.
\end{itemize}

\section{Related Work}
Self-supervised Learning has been a promising paradigm to learn useful image representations. Previous self-supervised methods try to design different handcrafted auxiliary tasks. Examples include context prediction~\cite{doersch2015unsupervised}, colorization~\cite{zhang2016colorful}, context encoder~\cite{pathak2016context}, jigsaw puzzle~\cite{noroozi2016unsupervised}, and rotation prediction ~\cite{gidaris2018unsupervised}, etc. 
Recently, contrastive learning has drawn much attention. Representative methods include Instance Discrimination~\cite{wu2018unsupervised}, MoCo~\cite{he2020momentum, chen2020improved}, and SimCLR~\cite{chen2020simple}. Contrastive methods learn an embedding space where features of different augmentations from the same image are attracted, and features of different images are separated. BYOL~\cite{grill2020bootstrap} and SimSiam~\cite{chen2021exploring} propose to abandon the negative samples and design some special techniques such as asymmetric architecture, momentum encoder and stop gradients to avoid the collapsed solution. Barlow Twins~\cite{zbontar2021barlow} and VICReg~\cite{bardes2021vicreg} propose to learn informative representations by reducing the redundancy or covariance of different dimensions. 

The clustering-based methods~\cite{caron2018deep, asano2019self, caron2019unsupervised, caron2020unsupervised, li2020prototypical, pmlr-v48-xieb16,yang2016joint} have also exhibited remarkable progress. They usually use a clustering tool to generate pseudo-labels for images and then classify the images with the generated pseudo-labels. The two processes alternate with each other.  DeepCluster~\cite{caron2018deep} uses the K-means algorithm to generate pseudo-labels for every epoch. SwAV~\cite{caron2020unsupervised} uses the Sinkhorn-Knopp algorithm \cite{asano2019self} to generate soft pseudo-labels and updates the pseudo-labels for every iteration. 
The most recent DINO~\cite{caron2021emerging} updates pseudo-targets using the output of the momentum teacher together with the sharpening and centering operations. Self-Classifier \cite{amrani2021self} proposes an end-to-end method to classify unlabeled datasets, which is closely related to \ourmethod, while \ourmethod uses a different way to prevent the collapsed solution problem. Detailed comparisons with Self-Classifier are shown in Sec \ref{sec:comp}. SCAN \cite{van2020scan} advocates a \textit{two-step approach} to mainly focus on unsupervised classification task.

The mutual information maximization between input and class predictions is a historical method. Bridle et al \cite{bridle1992unsupervised} first propose to utilize it as a clustering technique. IMSAT \cite{hu2017learning} uses it to encourage the predicted representations of augmented data to be close to those of the original data, and in the meantime regularizes the information dependency. IIC \cite{Ji_2019_ICCV} proposes a clustering objective to maximize the mutual information of \textit{different views}. Previous work of Deep InfoMax \cite{hjelm2018learning} proposes to learn representations by maximizing mutual information between \textit{different layers} of features.
The differences between \ourmethod and Deep Infomax are apparently: (1) The task of \ourmethod is to classify images by exploring their semantic relations, while the task of Deep InfoMax is to maximize the information between different neural layers. (2) \ourmethod measures the information between image and a discrete random variable, instead of high-dimensional continuous representation in Deep InfoMax. (3) We do not require the neural estimator \cite{belghazi2018mine} to estimate the mutual information.

\section{Method}

\begin{figure}
  \begin{center}
  \includegraphics[width=0.9\linewidth]{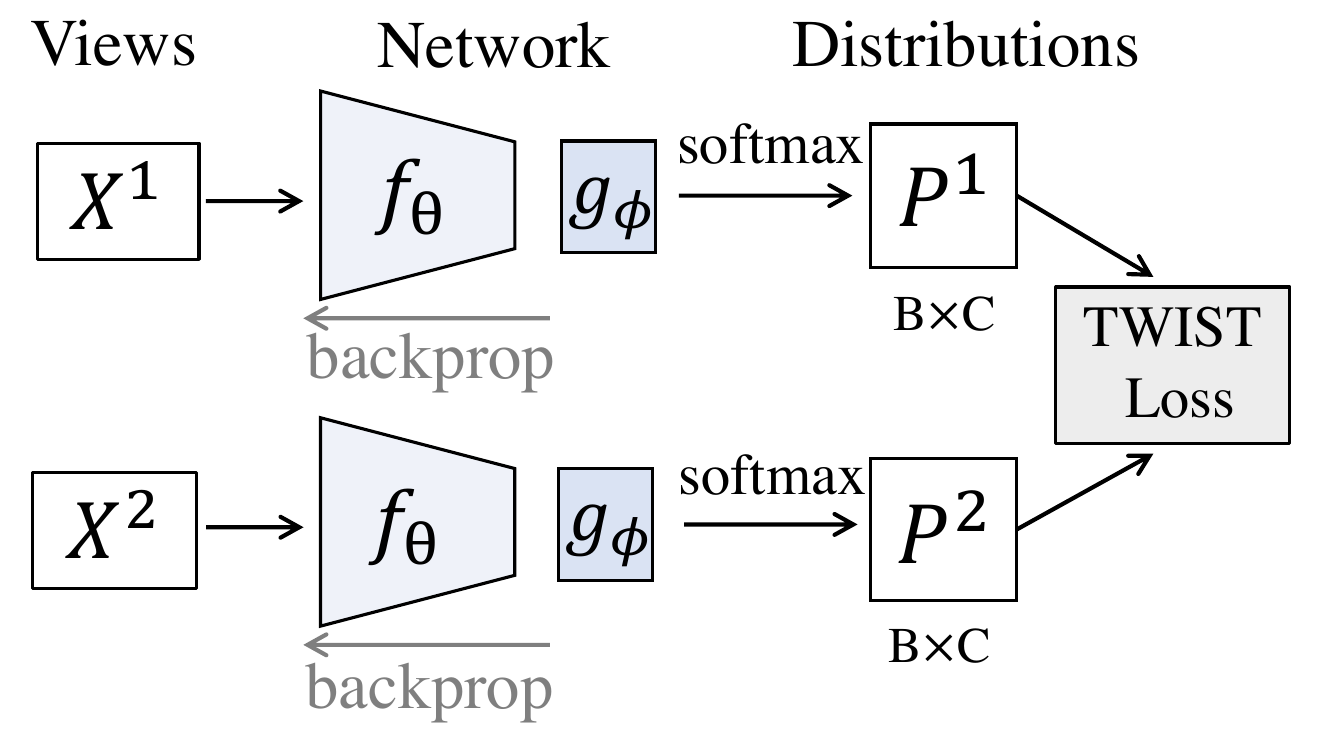}
  \end{center}
  \vspace{-0.4cm}
  \caption{Network architecture of \ourmethod.}
  \vspace{-0.4cm}
  \label{concept:main}
\end{figure}
Our goal is to learn an end-to-end unsupervised classification network to make accurate predictions and learn good representations. Without labels, we design the \ourmethod loss to make the predictions of two augmented images be recognized as the same class. In the meantime, \ourmethod loss regularizes the class distribution to make it sharp and diverse, which helps the network maximize the shared information between input images and output predictions. 

\subsection{Formulation}
\def\mycmd{0}
\if\mycmd0
Given an unlabeled dataset $\mathcal{S}$, we randomly sample a batch of images $\mathcal{X}\subset\mathcal{S}$ with a batch-size of $B$, and generate two augmented version $\mathcal{X}^{1}$ and $\mathcal{X}^{2}$ according to a predefined set of image augmentations. The two augmented images are fed into a siamese neural network composed of a backbone $f_\theta$ with parameters $\theta$, and a projection head terminated by a softmax operation $g_\phi$ with parameters $\phi$. The outputs of the neural networks $g_\phi \circ f_\theta$ are two probability distributions over $C$ categories $P^{k} = g_\phi(f_\theta(\mathcal{X}^{k}))$ ($k\in\{1,2\}$). The process is shown in Fig. \ref{concept:main}. 
Note that $P^{k} \in \mathbb{R}^{B \times C}$, and $P^{k}_i$ denotes the $i$-th row of $P^{k}$, i.e., probability distribution of the $i$-th sample.

With the probability distributions of two augmented views $P^{1}$ and $P^{2}$,
we define the learning objective as the symmetric form 
\def\formu{1}
\if\formu0
    \begin{equation}
        \small
        \mathcal{L}(P^{1}, P^{2})=\frac{1}{2}(\mathcal{L}(P^{1} || P^{2}) + \mathcal{L}(P^{2} || P^{1})),
        \label{form: sym}
    \end{equation}
    where $\mathcal{L}(P^{1} || P^{2}) = $
    \begin{equation}
        \small
        \underbrace{\frac{1}{B} \sum_{i=1}^{B} D_{KL}(P^{1}_{i}||P^{2}_{i})}_{\text{consistency term}} + \underbrace{\frac{\alpha}{B} \sum_{i=1}^B H(P^{1}_{i})}_{\text{sharpness term}}  - \beta\underbrace{H(\frac{1}{B} \sum_{i=1}^{B} P_{i}^{1})}_{\text{diversity term}},
        \label{form: kl}
    \end{equation}
and $\mathcal{L}(P^{2} || P^{1})$ is defined in the same way.
\else
    \begin{equation}
    \small
    \begin{aligned}
    \hspace{-0.2cm}
        \mathcal{L}(P^{1}, P^{2}) & = 
        \underbrace{\frac{1}{2B} \sum_{i=1}^{B} (D_{KL}(P^{1}_{i}||P^{2}_{i})+D_{KL}(P^{2}_{i}||P^{1}_{i}))}_{\text{consistency term}} \\ & + \frac{\alpha}{2} \sum_{k=1}^2 \underbrace{\frac{1}{B}\sum_{i=1}^B H(P^{k}_{i})}_{\text{sharpness term}}  - \frac{\beta}{2}\sum_{k=1}^2\underbrace{H(\frac{1}{B} \sum_{i=1}^{B} P_{i}^{k})}_{\text{diversity term}},
    \end{aligned}
    \label{form: kl}
    \end{equation}
\fi
where $D_{KL}(\cdot||\cdot)$ denotes the Kullback–Leibler divergence~\cite{kl_information} between two probability distributions. $H(\cdot)$ denotes the Entropy~\cite{entropy} of a specific probability distribution. $\alpha$ and $\beta$ are hyper-parameters to balance the two terms. 

Specifically, minimizing the consistency term makes the predictions of different views consistent, i.e., different augmentations of the same image are required to be recognized as the same class. For the sharpness term, we minimize the entropy of class distribution for each sample to regularize the output distribution to be sharp, which makes each sample have a deterministic assignment (i.e., one-hot vector in the ideal case). Besides, features of samples assigned to the same category will be more compact. For the diversity term, we try to make the predictions for different samples be diversely distributed to avoid the network assigning all images to the same class. This is achieved by maximizing the entropy of the mean distribution across different samples $H(\frac{1}{B} \sum_{i=1}^{B} P_{i})$.

\else

done

\fi

\subsection{Theoretical Explanation}
In this part, we explain \ourmethod loss as an objective that optimizes two things: (1) minimizing the class prediction disagreements between augmentations through the consistency term. (2) maximizing the mutual information between the input image and the output class probability distributions, which is achieved by the sharpness term and diversity term. The mutual information between a random variable $X$ representing input image and the predicted class $Y$ is
\begin{equation}
\small
\begin{aligned}
\hspace{-0.2cm}
     - &I(X,Y) = H(Y|X) - H(Y)\\
           & = \mathop{\mathbb{E}}\limits_x \left[- \sum_{y} p(y|x){\rm log} p(y|x) \right] - H(Y)\\
           & \approx \underbrace{-\frac{1}{|\mathcal{X}|}\sum_{x \in \mathcal{X}}\sum_{y} p(y|x){\rm log} p(y|x)}_{\text{sharpness term}}  - \underbrace{H(\frac{1}{|\mathcal{X}|} \sum_{x \in \mathcal{X}} p(Y|x))}_{\text{diversity term}},
\end{aligned}
\label{form:info}
\end{equation}
where $H(Y|X)$ is the Conditional Entropy. 
The approximation is derived from the Monte Carlo estimation of the expectation on $x$. From Eq. \ref{form:info}, we observe that the first term is the sharpness term, and the second term is the diversity term. The Monte Carlo estimation of the diversity term is because of the following relation 
\begin{equation}
\begin{aligned}
    p(Y) & = \int_x p(Y|x)p(x) dx 
         \approx \frac{1}{|\mathcal{X}|} \sum_{x \in \mathcal{X}} p(Y|x).
\end{aligned}
\end{equation}
In this perspective, the \ourmethod loss can be interpreted as minimizing the disagreement of different augmentations and simultaneously maximizing the mutual information between the input image and output class predictions, when $\alpha = \beta = 1$. From this view, we show the explanation why the \ourmethod loss can avoid the collapsed solutions. Specifically, \ourmethod loss requires the output class prediction $Y$ to preserve information with the input image $X$ as much as possible. When all images are predicted to the same class, the information shared with the input image will vanish.

\subsection{Amplifying Variance for Better Optimization}
In practice, directly optimizing the \ourmethod loss will derive sub-optimal solutions. Specifically, we find that the consistency term and the sharpness term are easy to minimize while the diversity term is difficult to maximize. We visualize the standard deviations of each row and each column of the features before softmax, as illustrated in Fig.~\ref{bnid}. The column standard deviation keeps small during the training process, which makes the probability of each class tend to be similar across different samples in a mini-batch. This will cause the low diversity of classification results. To solve the problem, we add a batch normalization \cite{ioffe2015batch} before the softmax to amplify the variance of probabilities in each class to force them to be separated. 
By adding the batch normalization layer, the diversity term is well optimized. 
The batch normalization before softmax brings about 5\% improvements in ImageNet linear classification. More analyses and experiments are shown in the Ablation section. 

\section{Methodology Comparisons}\label{sec:comp}
In this section, we discuss the comparisons of \ourmethod to the most related works in details.

\textbf{DINO \cite{caron2021emerging}} utilizes the class distributions of the momentum encoder to form a self-distillation process, and achieves good performance. \ourmethod shares similarity with DINO for that \ourmethod also learns the class probability distributions. However, \ourmethod is different from DINO in the following aspects: \textbf{(1)} different objectives, \textbf{(2)} different mechanisms to avoid collapsed solutions, and \textbf{(3)} different training manners.
We give the objective function of DINO as 
\begin{equation}
\small
\begin{aligned}
    L_{DINO} &= CE({\rm \textbf{stop}}(P_t(x_1)), P_s(x_2))\\
             &= D_{KL}({\rm \textbf{stop}}(P_t(x_1))||P_s(x_2)) + \bcancel{H({\rm \textbf{stop}}(P_t(x_1)))},
\end{aligned}
\label{form:dino}
\end{equation}
where \textbf{stop} means the gradients is not back-propagated.
Here we re-write the DINO loss by the sum of KL-divergence plus a sample-wise entropy term, which we intend to make a clear comparison with \ourmethod loss in Eq. \ref{form: kl}. 
We see that DINO loss is equal to the KL-divergence term in \ourmethod, without giving the loss-guided constrain to the sample-wise entropy (the slash term). In contrast, \ourmethod minimizes the entropy of each sample and simultaneously maximizes the diversity.

DINO incorporates sharpening and centering operations to avoid assigning all samples to the same class, and it relies on momentum encoder to enable training.
\ourmethod can work naturally with the unified loss function, eliminating the reliance on momentum encoder. \ourmethod uses a more explainable and straightforward way to avoid collapsed solutions. 
Empirically, \ourmethod performs much better at semi-supervised settings and fine-tuning settings while achieving competitive results with DINO in the linear setting.

\textbf{SwAV \cite{caron2020unsupervised}} uses the Sinkhorn-Knopp algorithm \cite{asano2019self} to generate soft pseudo-labels for samples in a mini-batch. The differences between \ourmethod and SwAV are distinct. \ourmethod does not rely on the Sinkhorn-Knopp algorithm or any process to generate pseudo labels. Instead, the \ourmethod loss naturally helps the network learn meaningful assignments and representations.

\textbf{Self-Classifier \cite{amrani2021self}} is an end-to-end self-supervised method that designs a classification network without relying on momentum-encoder or clustering techniques. 
The main differences between \ourmethod and Self-Classifier lie in the loss function. The loss function of self-classifier is $L_{SC}(x_1, x_2) = - \sum_{y=1}^{C} p(x_2|y)log(p(y|x_1))$.

Training the loss function is equal to optimizing the cross entropy between $p(y|x_1)$ and $p(y|x_2)$, with the assumption that $p(y)$ is a \textit{uniform distribution}. The calculation about $p(x_2|y)$ include the softmax operation across samples in a mini-batch, which can be regarded as setting the iterations of Sinkhorn-Knopp algorithm to 1~\cite{caron2021emerging}. 
In contrast, our \ourmethod loss can measure and maximize the mutual information between $x$ and $y$, making the method explainable. Moreover, \ourmethod performs much better than Self-Classifier on all benchmarks.

\textbf{Barlow Twins \cite{zbontar2021barlow}} proposes to learn representations by decorrelating different feature dimensions. \ourmethod shares the clean architectures as Barlow Twins. However, the learning objective and principle are quite different. We aim at learning an unsupervised classifier by our proposed loss function, while the objective of Barlow Twins is to minimize the redundancy between feature dimensions.

\section{Main Results}
We evaluate the performances of \ourmethod on ImageNet un/semi-supervised classification, transfer learning, and a wide range of downstream tasks. We set $C=4096$ for the
representation learning
and $C=1000$ for the ImageNet unsupervised classification in accordance with the standard ImageNet class number. We adopt the multi-crop \cite{caron2020unsupervised} augmentation strategy and also report performances without multi-crop for fair comparisons. We set $\alpha=\beta=1$ for ResNet and $\alpha=0.4, \beta=1$ for ViT. More implementation details can be found in Appendix. For all downstream tasks, we strictly follow the common evaluation procedures. All \ourmethod models are trained on the train set of ImageNet ILSVRC-2012 which has $\sim$1.28 million images \cite{deng2009imagenet}. 

\subsection{Training Strategy}

\textbf{Self-labeling for ResNets:} For ResNets, the multi-crop strategy helps improve the linear classification performance from 72.6\% to 74.3\%, shown in Tab.~\ref{lc_other}. However, compared with the performance improvement of SwAV (from 71.8\% to 75.3\%), the performance gain of \ourmethod is much smaller (3.5\% v.s. 1.7\%). Such phenomenon has also been observed by~\cite{caron2021emerging}. 
Specifically, SwAV uses the global crops to generate relatively accurate pseudo-labels as supervision to train the local crops. However, in our method, the global crops and local crops are regarded equally. Thus the noisy local crops can also affect the accurate predictions of global crops, which will not happen in methods of SwAV and DINO. To take full advantage of the multi-crop strategy, we add a self-labeling stage after the regular training. Specifically, we use the outputs of the global crops as supervision to train other crops. Different with SwAV: \textbf{(1)} we directly use our network outputs as supervision, instead of the outputs of the Sinkhorn-Knopp algorithm, \textbf{(2)} we only use the samples in a mini-batch whose confidences surpass a predefined threshold. With only 50 epochs of self-labeling after finishing the regular training, we have another 1.2\% performance gains (from 74.3\% to 75.5\%). We empirically show that the proposed self-labeling could not improve the performance of SwAV~\cite{caron2020unsupervised} or DINO~\cite{caron2021emerging}. 

\textbf{Momentum Encoder for ViT:} For Vision Transformers \cite{dosovitskiy2020image,touvron2021training}, we do not use the self-labeling process. Instead, we adopt the momentum encoder design, which is widely adopted to train ViT-based models \cite{caron2021emerging, chen2021empirical}. Specifically, one tower of our siamese network is updated by the exponential moving average of the parameters from the other tower, similar as \cite{he2020momentum} and \cite{grill2020bootstrap}. The whole network is updated by the \ourmethod loss. Although we use the momentum encoder as the default setting for ViT backbones, \ourmethod using ViT as backbone can also work without it and achieves 72.5\% ImageNet Top-1 linear accuracy for Deit-S 300 epochs. 
The momentum encoder is only for accuracy gains. We do not use it for CNNs.

\newcommand{\cmark}{\ding{51}}
\newcommand{\xmark}{\ding{55}}
\definecolor{backcolor}{RGB}{232, 242, 255}

\begin{table}[h]
\begin{center}
\small
\begin{tabular}{l>{\centering\arraybackslash}p{0.6cm}>{\centering\arraybackslash}p{0.6cm}>{\centering\arraybackslash}p{0.6cm}>{\centering\arraybackslash}p{0.6cm}>{\centering\arraybackslash}p{0.6cm}>{\centering\arraybackslash}p{0.6cm}}
\Xhline{2\arrayrulewidth}
\rule{0pt}{1.1\normalbaselineskip}
\multirow{2}{*}{\textbf{Method}} & \multicolumn{2}{c}{\textbf{1\%} Labels} & \multicolumn{2}{c}{\textbf{10\%} Labels} & \multicolumn{2}{c}{\textbf{100\%} Labels}\\
 & Top1 & Top5 & Top1 & Top5 & Top1 & Top5\\
\Xhline{2\arrayrulewidth}
\multicolumn{7}{l}{\textit{\textbf{ResNet-50}}} \rule{0pt}{2.6ex}\\
SUP    & 25.4 & 48.4 & 56.4 & 80.4 & 76.5 & - \\
SimCLR & 48.3 & 75.5 & 65.6 & 87.8 & 76.5 & 93.5 \\
BYOL   & 53.2 & 78.4 & 68.8 & 89.0 & 77.7 & 93.9\\
SwAV   & 53.9 & 78.5 & 70.2 & 89.9 & - & -\\
DINO   &  52.2  &  78.2   &  68.2   &  89.1   & - & -\\
BarlowTwins & 55.0 & 79.2 & 69.7 & 89.3 & - & -\\
\rowcolor{backcolor} \ourmethod & \bf 61.2 & \bf 84.2 & \bf 71.7 & \bf 91.0 & \bf 78.4 & \bf 94.6\\
\Xhline{2\arrayrulewidth}
\multicolumn{7}{l}{\textit{\textbf{ResNet-50$\times$2}}} \rule{0pt}{2.6ex}\\
SimCLR & 58.5 & 83.0 & 71.7 & 91.2 & - & -\\
BYOL & 62.2 & 84.1 & 73.5 & 91.7 & - & -\\
\rowcolor{backcolor} \ourmethod & \bf 67.2& \bf 88.2 & \bf 75.3 & \bf 92.8 & \bf 80.3 & \bf 95.4\\
\Xhline{2\arrayrulewidth}
\multicolumn{7}{l}{\textit{\textbf{ViT-B/16}}}\rule{0pt}{2.6ex}\\
DINO & 67.3 & 88.2 & 74.6 & 92.0 & \bf 82.8 & - \\
\rowcolor{backcolor} \ourmethod & \bf 69.6 & \bf 89.7 & \bf 76.5 & \bf 93.1 & \bf 82.8 & \bf 96.3 \\
\Xhline{2\arrayrulewidth}
\end{tabular}
\end{center}
\vspace{-0.4cm}
\caption{Semi-supervised classification results on ImageNet. We report top-1 and top-5 center-crop accuracies, from 1\% to 100\%.}
\label{semi}
\vspace{-0.4cm}
\end{table}

\subsection{Semi/Fully-supervised Fine-tuning} We fine-tune the pre-trained \ourmethod model on a subset of ImageNet, following the standard procedure~\cite{chen2020simple, grill2020bootstrap}. From Tab.~\ref{semi}, we see that \ourmethod outperforms all other state-of-the-art methods by large margins. With only $1\%$ of labeled data, \ourmethod achieves $61.2\%$ top-1 accuracy with ResNet-50, surpassing the previous best result by $6.2\%$. The trend is preserved when the backbone becomes larger. For example, \ourmethod achieves $67.2\%$ ($+5.0\%$) top-1 accuracy with ResNet-50w2. With $10\%$ of labeled data, \ourmethod still achieves the best result. 

We also fine-tune the pre-trained \ourmethod model on the full ImageNet. The results are shown in the last two columns of Tab.~\ref{semi}. With our pre-trained model as initialization, ResNet-50 achieves $78.4\%$ top-1 accuracy, surpassing the model trained from scratch by a large margin ($+1.9\%$).

\begin{table}[h]
\begin{center}
\begin{tabular}{l|cccc}
Method & NMI & ARI & AMI & ACC\\
\Xhline{2\arrayrulewidth}
SCAN   & 72.0 & 27.5 & 51.2 & 39.9\\
SeLa        & 65.7 & 16.2 & 42.0 & -  \\
SelfClassifier  & 64.7 & 13.2 & 46.2 & - \\
\rowcolor{backcolor} \ourmethod  & \bf 74.3 & \bf 30.0 & \bf 57.7 & \bf 40.6 \\
\Xhline{2\arrayrulewidth}
\end{tabular}
\end{center}
\vspace{-0.4cm}
\caption{Unsupervised classification results on ImageNet. All numbers are reported on the validation set of ImageNet. Comparison methods include SCAN \cite{van2020scan}, SeLa \cite{asano2019self}, and Self Classifier \cite{amrani2021self}.}
\label{unsup_class}
\vspace{-0.2cm}
\end{table}
\textbf{Unsupervised Classification:} Finally, we evaluate the unsupervised classification results using no label as supervision. For evaluation, the outputs of \ourmethod are directly mapped to the real labels by the Kuhn–Munkres \cite{kuhn1955hungarian} algorithm. Tab. \ref{unsup_class} shows the results. \ourmethod with ResNet-50 backbone outperforms previous best results by $2.3\%$ NMI. Details are shown in appendix.

\begin{table}[h]
\small
        \scalebox{0.97}{
        \begin{tabular}{lccccc}
        \Xhline{2\arrayrulewidth}
        \bf Method  & \bf Network & \bf Param & \bf Epoch & \bf Top1 & \bf Top5 \rule{0pt}{2.6ex}\\
        \Xhline{2\arrayrulewidth}
        \multicolumn{6}{l}{\textit{\textbf{ResNet-50 without multi-crop}}}\rule{0pt}{2.6ex}\\
        MoCo v2 & RN50 & 24M & 800 & 71.1 & 90.1 \\
        SimCLR  & RN50 & 24M & 1000 & 69.3 & 89.0 \\
        BarlowTwins & RN50 & 24M & 1000 & 73.2 & 91.0 \\
        BYOL    & RN50 & 24M & 1000 & \textbf{74.3} & \textbf{91.6} \\
        SelfClassifier & RN50 & 24M & 800 & 69.7 & 89.3 \\
        SwAV    & RN50 & 24M & 800  & 71.8 & - \\
        \rowcolor{backcolor} \ourmethod & RN50 & 24M & 800 & 72.6 & 91.0 \\
        \multicolumn{6}{l}{\textit{\textbf{ResNet-50 with multi-crop}}}\\
        SwAV    & RN50 & 24M & 800 & 75.3 & - \\
        DINO    & RN50 & 24M & 800 & 75.3 & 92.5 \\
        \rowcolor{backcolor} \ourmethod    & RN50 & 24M & 300 & 75.0 & 92.4 \\
        \rowcolor{backcolor} \ourmethod    & RN50 & 24M & 800 & \textbf{75.5} & \textbf{92.5} \\
        \Xhline{2\arrayrulewidth}

        \multicolumn{6}{l}{\textit{\textbf{Wider ResNet}}} \rule{0pt}{2.6ex}\\
        SimCLR  & RN50w2 & 94M & 1000 & 74.2 & 92.0 \\
        CMC    & RN50w2 & 94M & -   & 70.6 & 89.7 \\
        SwAV   & RN50w2 & 94M & 800 & 77.3 & - \\
        BYOL   & RN50w2 & 94M & 1000 & 77.4 & 93.6\\
        \rowcolor{backcolor} \ourmethod    & RN50w2 & 94M & 300 & \bf 77.7 & \bf 93.9 \\
        \Xhline{2\arrayrulewidth}
        \multicolumn{6}{l}{\textit{\textbf{Vision Transformer}}} \rule{0pt}{2.6ex}\\
        MoCo-v3  & Deit-S/16          & 21M & 300 & 72.5 & - \\
        DINO  & Deit-S/16          & 21M & 300 & 75.9 & - \\
        \rowcolor{backcolor} \ourmethod     & Deit-S/16          & 21M & 300 & \bf 76.3 & \bf 92.7 \\
        MoCo-v3  & ViT-B/16 & 86M & 300 & 76.5 & - \\
        DINO  & ViT-B/16 & 86M & 800 & 78.2 & \bf 93.9 \\
        \rowcolor{backcolor} \ourmethod     & ViT-B/16          & 86M & 300 & \bf 78.4 & 93.8 \\
        \Xhline{2\arrayrulewidth}
        \end{tabular}}
    \vspace{-0.1cm}
    \caption{Linear classification results. We report and compare results with different backbones. 
    }
        \label{lc_in}
        \vspace{-0.4cm}
\end{table}

\subsection{Linear Classification} We evaluate the performance of linear classification on ImageNet. Specifically, we add a linear classifier on top of the frozen backbone network and measure the top-1 and top-5 center-crop classification accuracies, following previous works \cite{zhang2016colorful, he2020momentum, chen2020simple}. The results are shown in Tab.~\ref{lc_in}. \ourmethod outperforms other state-of-the-art methods, achieving 75.5\% top-1 accuracy on ResNet-50. Besides, we also train \ourmethod with other backbones of neural networks, including wider ResNet-50 and Vision Transformers. For wider ResNet-50, the superiority of \ourmethod becomes more apparent. \ourmethod outperforms SwAV and BYOL by $0.4\%$ and $0.3\%$ respectively using ResNet-50w2 and 300 training epochs. For Vision Transformers, \ourmethod is also comparable with other state-of-the-art methods, achieving 78.3\% Top-1 accuracy with 300 epochs. Besides, we find that ViT is very sensitive to hyper-parameters and training it is very costly, we believe better results can be achieved after well-tuning.

\definecolor{ccolor}{RGB}{250,235,215}
\definecolor{dcolor}{RGB}{155,226,235}
\begin{table*}[t]
\begin{center}
\small
\scalebox{0.95}{
\begin{tabular}{lcccccccccccc}
\Xhline{2\arrayrulewidth}
Method  & Food & Cifar10 & Cifar100 & Sun397 & Cars & Aircraft & VOC & DTD & Pets & Caltech & Flowers & Avg \rule{0pt}{2.6ex}\\
\Xhline{2\arrayrulewidth}
\multicolumn{12}{l}{\textit{\textbf{Linear evaluation:}}}& \rule{0pt}{2.6ex}\\
SimCLR & 68.4 & 90.6 & 71.6 & 58.8 & 50.3 & 50.3 & 80.5 & 74.5 & 83.6 & 90.3 & 91.2 & 73.6 \\
BYOL & 75.3 & 91.3 & \textbf{78.4} & 62.2 & \textbf{67.8} & 60.6 & 82.5 & 75.5 & 90.4 & 94.2 & \textbf{96.1} & \bf 79.5 \\
SUP & 72.3 & \textbf{93.6} & 78.3 & 61.9 & 66.7 & \textbf{61.0} & 82.8 & 74.9 & 91.5 & \textbf{94.5} & 94.7 & 79.3\\
\rowcolor{backcolor} \ourmethod & \textbf{78.0} & 91.2 & 74.4 & \textbf{66.8} & 55.2 & 53.6 & \textbf{85.7} & \textbf{76.6} & \bf 91.6 & 91.1 & 93.4 & 78.0\\
\Xhline{2\arrayrulewidth}
\multicolumn{12}{l}{\textit{\textbf{Fine-tune:}}}& \rule{0pt}{2.6ex}\\
Random & 86.9 & 95.9 & 80.2 & 53.6 & 91.4 & 85.9 & 67.3 & 64.8 & 81.5 & 72.6 & 92.0 & 79.3 \\
SimCLR & 87.5 & 97.4 & 85.3 & 63.9 & 91.4 & 87.6 & 84.5 & 75.4 & 89.4 & 91.7 & 96.6 & 86.4\\
BYOL & 88.5 & 97.8 & 86.1 & 63.7 & 91.6 & \textbf{88.1} & 85.4 & 76.2 & 91.7 & \textbf{93.8} & 97.0 & 87.3\\
SUP & 88.3 & 97.5 & 86.4 & 64.3 & \textbf{92.1} & 86.0 & 85.0 & 74.6 & 92.1 & 93.3 & \bf 97.6 &87.0\\
\rowcolor{backcolor} \ourmethod & \bf 89.3 & \bf 97.9 & \bf 86.5 & \textbf{67.4} & 91.9 & 85.7 & \textbf{86.5} & \bf 76.4 & \textbf {94.5} & 93.5 & 97.1 & \bf 87.9\\
\Xhline{2\arrayrulewidth}
\end{tabular}
}
\vspace{-0.1cm}
\caption{Transfer learning results on eleven datasets, including linear evaluation and fine-tuneing. We use ResNet-50 as backbone and pre-trained on ImageNet. We calculate the average of performanes on these datasets and report it at the last column. \ourmethod performs best on the fine-tuning setting, which is in accordance to the advantage on the semi-supervised fine-tuning setting.}
\vspace{-0.3cm}
\label{transfer}
\end{center}
\end{table*}

\subsection{Transfer Learning} To further validate the features learned by \ourmethod, we evaluate \ourmethod model on eleven different datasets, including Food101 \cite{bossard14}, CIFAR10, CIFAR100 \cite{krizhevsky2009learning}, SUN397 \cite{xiao2010sun}, Cars \cite{Krause2013CollectingAL}, FGVC-Aircraft \cite{maji13fine-grained}, Pascal VOC2007 \cite{Everingham2009ThePV}, Describable Textures Dataset (DTD) \cite{cimpoi14describing}, Oxford-IIIT Pet \cite{Parkhi2012CatsAD}, Caltech101 \cite{Li2004LearningGV}, and Flowers \cite{nilsback2008automated}. Tab.~\ref{transfer} shows the results. We also report the results of linear classification models for a comprehensive comparison. \ourmethod performs competitively on these datasets. In some datasets, \ourmethod  achieves improvements over $1\%$. \ourmethod outperforms the supervised models on seven out of eleven datasets. We also observe that our method shows more advantages over other methods in the fine-tune setting.

\subsection{Detection and Segmentation} We evaluate the learned representations of \ourmethod on object detection and instance segmentation. We conduct experiments on Pascal VOC \cite{Everingham2009ThePV} and MS COCO \cite{lin2014microsoft}.
We use ResNet-50 with Feature Pyramid Network (FPN)~\cite{lin2017feature} as the backbone architecture. For Pascal VOC, we use the Faster R-CNN \cite{ren2015faster} as the detector. For MSCOCO, we follow the common practice to use the Mask R-CNN \cite{he2017mask}. In implementation, we use Detectron2 \cite{wu2019detectron2}, with the same configurations as
~\cite{tian2020makes} and~\cite{wang2020DenseCL}. 
Tab.~\ref{det} shows the results. We can see that \ourmethod performs better on all three tasks, demonstrating the advantages of using \ourmethod as the pre-trained model in object detection and instance segmentation. Besides, we find that the FPN architecture is important for the category-level self-supervised learning methods to achieve good performance. Analysis is given in Appendix.

\begin{table*}[t]
    \small
    \begin{center}
    \scalebox{0.95}{
    \begin{minipage}{0.7\linewidth}
        \begin{center}
        \begin{tabular}{l|ccc|ccc|ccc}
        \Xhline{2\arrayrulewidth}
        \rule{0pt}{1.1\normalbaselineskip}
        \multirow{2}{*}{Method}  & \multicolumn{3}{c|}{VOC07+12 detection} & \multicolumn{3}{c|}{COCO detection} & \multicolumn{3}{c}{COCO instance seg} \\
         & AP$_{\rm all}$ & AP$_{\rm 50}$ & AP$_{\rm 75}$ & AP$^{\rm bb}_{\rm all}$ & AP$^{\rm bb}_{\rm 50}$ & AP$^{\rm bb}_{\rm 75}$ & AP$^{\rm mk}_{\rm all}$ & AP$^{\rm mk}_{\rm 50}$ & AP$^{\rm mk}_{\rm 75}$ \rule[-0.9ex]{0pt}{0pt}\\
        \Xhline{2\arrayrulewidth}
        Moco-v2 & 56.4 & 81.6 & 62.4 & 39.8 & 59.8 & 43.6 & 36.1 & 56.9 & 38.7 \rule{0pt}{2.6ex}\\
        SimCLR $^\dagger$& \bf 58.2 & 83.8 & 65.1 & 41.6 & 61.8 & 45.6 & 37.6 & 59.0 & 40.5 \\
        SwAV & 57.2 & 83.5 & 64.5 & 41.6 & 62.3 & \bf 45.7 & \bf 37.9 & 59.3 & \bf 40.8 \\
        DC-v2 $^\dagger$& 57.0 & 83.7 & 64.1 & 41.0 & 61.8 & 45.1 & 37.3 & 58.7 & 39.9 \\
        DINO $^\dagger$& 57.2 & 83.5 & 63.7 & 41.4 & 62.2 & 45.3 & 37.5 & 58.8 & 40.2 \\
        DenseCL & 56.9 & 82.0 & 63.0 & 40.3 & 59.9 & 44.3 & 36.4 & 57.0 & 39.2 \\
        \rowcolor{backcolor} \ourmethod & 58.1 & \bf 84.2 & \bf 65.4 & \bf 41.9 & \bf 62.6 & \bf 45.7 & \bf 37.9 & \bf 59.7 & 40.6\\
        \Xhline{2\arrayrulewidth}
        \end{tabular}
        \end{center}
        \vspace{-0.5cm}
        \caption{Detection and instance segmentation. $^\dagger$means that we download the pre-trained models and conduct the experiments. For the VOC dataset, we run five trials and report the average. The performance is measured by Average Precision (AP). DC-v2 denotes the DeepCluster-v2.}
        \label{det}
    \end{minipage}
    \hspace{0.005\linewidth}
    \begin{minipage}{0.27\linewidth}
        \begin{center}
        \begin{tabular}{l|cc}
        \Xhline{2\arrayrulewidth}
        \rule{0pt}{1.1\normalbaselineskip}
        \multirow{2}{*}{Method}  & \multicolumn{2}{c}{FCN-FPN} \\
         & VOC & Cityscapes  \\
        \Xhline{2\arrayrulewidth}
        Sup & 67.7 & \bf 75.4  \rule{0pt}{2.6ex}\\
        Moco-v2 & 67.5 & \bf 75.4 \\
        SimCLR  & \underline{72.8} & \underline{74.9} \\
        SwAV & 71.9 & 74.4 \\
        DC-v2 & 72.1 & 73.8 \\
        DINO & 71.9 & 73.8 \\
        \rowcolor{backcolor} \ourmethod & \bf 73.3 & 74.6\\
        \Xhline{2\arrayrulewidth}
        \end{tabular}
        \end{center}
        \vspace{-0.5cm}
        \caption{Results of semantic segmentation with FCN-FPN backbone. All results are averaged over five trials.}
        \label{ss}
    \end{minipage}
    }
    \end{center}
    \vspace{-0.7cm}
\end{table*}

We also evaluate \ourmethod on semantic segmentation, using FCN \cite{long2015fully} as architectures. We use the MMSegmentation \cite{mmseg2020} to train the architectures. Tab.~\ref{ss} shows the results on Pascal VOC \cite{Everingham2009ThePV} and Cityscapes \cite{cordts2016cityscapes}. The results indicate that \ourmethod is competitive to other state-of-the-art methods.

\section{Ablation Study}

\begin{figure*}[h]
    \vspace{-0.4cm}
   \begin{center}
   \includegraphics[width=0.95\linewidth]{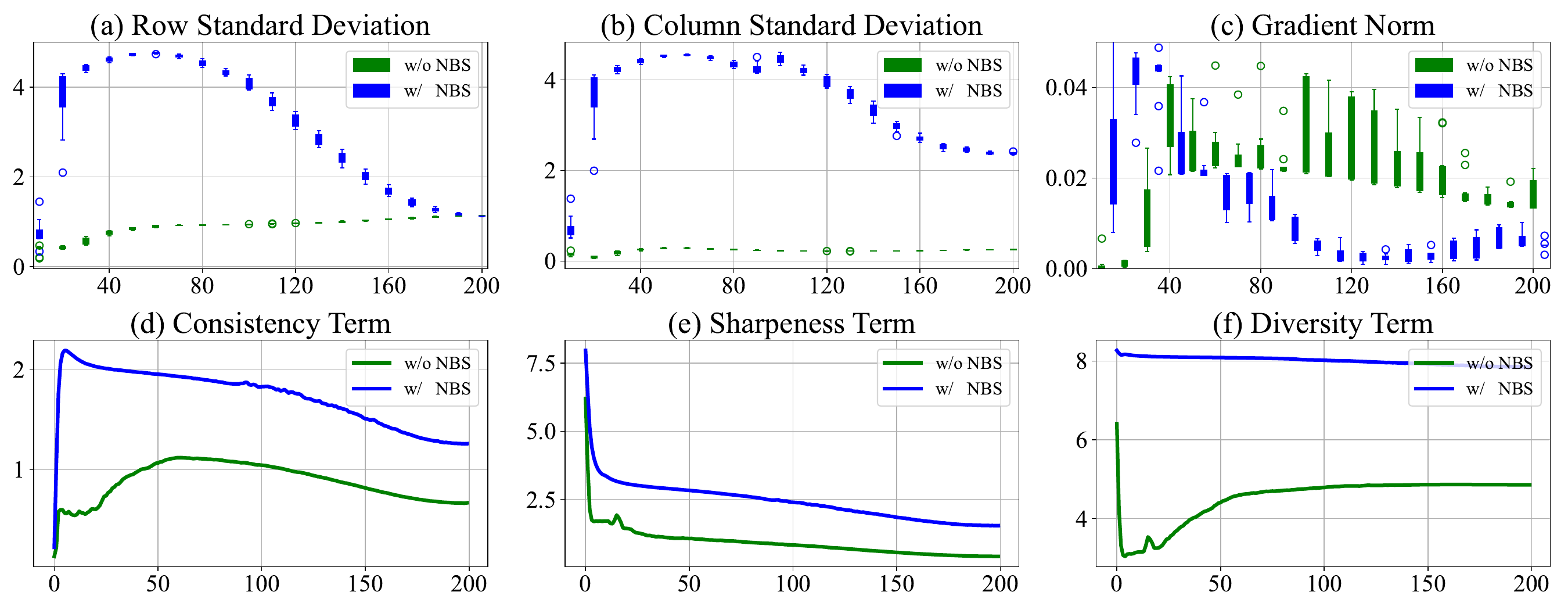}
   \end{center}
   \vspace{-0.7cm}
   \caption{We show the statistical characteristics of the output before softmax operation with and without NBS and the training curves.
   }
   \vspace{-0.4cm}
   \label{bnid}
\end{figure*}

\textbf{Amplifying Variance before Softmax:} We study the amplifying variance operation of the batch \underline{n}ormalization \underline{b}efore \underline{s}oftmax (abbreviated as NBS). 
From the loss values in Table \ref{nbs}, we can see that the model with NBS is optimized much better than the model without NBS. NBS brings 5.1\% top-1 accuracy improvement and 12\% NMI improvement on ImageNet, demonstrating the effectiveness of NBS. 
To better understand how NBS works, we look at the behaviors of the models with and without NBS. Fig. \ref{bnid} (a) and (b) show the row and column standard deviations of the output before softmax (input of BN for the NBS model). Although the intermediate processes are different, the row standard deviations are closing when training is finished. For the column deviations, it is not the case. The column standard deviation of NBS model is much larger than the model without NBS at the end of the training, indicating that the samples in the same batch tend to give similar predictions. This is also reflected in Fig. \ref{bnid} (f), from which we see that the diversity term of the model with NBS is better optimized than the model without NBS.
The observation indicates that the model without NBS tends to be column-collapsed. Although the solution is not fully collapsed, it tends to output similar predictions for samples in a mini-batch. NBS can successfully avoid the degenerated solution because batch normalization will force the standard deviation of each column to be one. Fig. \ref{bnid} (c) shows the magnitude and stability of the gradients from the optimization perspective.
\begin{table}[h]
    \centering
    \scalebox{0.92}{
    \begin{tabular}{c|ccc|cc}
          NBS   & Loss & ACC & NMI & std$_c$ & std$_r$ \rule{0pt}{2.6ex}\\
        \Xhline{2\arrayrulewidth}
         \cmark   & -5.05 & 70.6 & 59.0 & 2.37 & 1.12 \rule{0pt}{2.6ex}\\
         \xmark & -3.78 & 65.5 & 47.0 & 0.26 & 1.14 \\
    \end{tabular}}
    \vspace{-0.2cm}
    \caption{Ablation study on batch normalization before softmax.}
    \label{nbs}
    \vspace{-0.3cm}
\end{table}

\definecolor{LightCyan}{rgb}{0.88,1,1}
\definecolor{Gray}{gray}{0.85}
\newcolumntype{a}{>{\columncolor{Gray}}c}
\begin{table}[t]
    \begin{center}
    \begin{tabular}{cc|cc|cc}
    $L_{s}$ & $L_{d}$ & $L_{s} =$ & $L_{d}$ = & $|g|$ & acc \rule{0pt}{2.6ex}\\
    \Xhline{2\arrayrulewidth}
    \xmark & \xmark & 8.28 & 8.28 & 0 & 0.1 \rule{0pt}{2.6ex}\\
    \xmark & \cmark & 8.27 & 8.28 & 0 & 0.1 \\
    \cmark & \xmark & 2.59 & 6.42 & 0.01 & 56.1 \\
    \cmark & \cmark & 1.51 & 7.87 & 0.02 & 70.9\\
    \end{tabular}
    \end{center}
    \vspace{-0.5cm}
    \caption{Ablation study on the loss terms. Here $L_s$ and $L_d$ denote the sharpness and diversity term respectively. $|g|$ denotes the mean magnitude of gradients before the last batch normalization and ``acc'' is the linear accuracy. Models are trained for 50 epochs.}
    \label{ab_component}
    \vspace{-0.4cm}
\end{table}
\begin{table}[t]
    \begin{center}
    \begin{tabular}{l|cccc}
    multi-crop  &   \cmark & \cmark & \xmark & \xmark \rule{0pt}{2.6ex}\\
    
    self-labeling & \cmark & \xmark & \cmark & \xmark  \rule{0pt}{2.6ex}\\
    \Xhline{2\arrayrulewidth}
    acc &           75.5       & 74.0 & 73.8 & 72.6 \rule{0pt}{2.6ex}\\
    \end{tabular}
    \end{center}
    \vspace{-0.5cm}
    \caption{Ablation study on multi-crop and self-labeling. We report the linear accuracy. Models are trained for 800 epochs.}
    \label{lc_other}
    \vspace{-0.6cm}
\end{table}

\textbf{Impact of Loss Terms:} 
We test the impact of the loss terms in \ourmethod. As shown in Tab.~\ref{ab_component}, the models trained without the sharpness term generate collapsed solutions: The entropy for each sample is as large as the entropy of a uniform distribution, and the gradient magnitude rapidly decreases to $0$. In contrast, the models trained without the diversity term do not generate collapsed solutions, but their performances deteriorate significantly. Theoretically, models trained without the diversity term will also lead to collapsed solutions, i.e., outputting the same one-hot distributions. However, the batch-normalization before the softmax operation helps avoid the problem because it can separate the probabilities in different columns and force them to have a unit standard deviation. Therefore, all three terms are indispensable for \ourmethod.

\textbf{Multi-crop and Self-labeling:} Tab.~\ref{lc_other} shows the ablation results on multi-crop and self-labeling, where the models are trained for 800 epochs. We observe that the multi-crop and self-labeling can improve the performance respectively, and the best result comes from the combination of both.

\definecolor{mygreen}{RGB}{69, 139, 116}
\definecolor{mygreen2}{RGB}{0, 139, 0}
\begin{figure}[h]
        \vspace{-0.1cm}
        \begin{center}
        \includegraphics[width=0.9\linewidth]{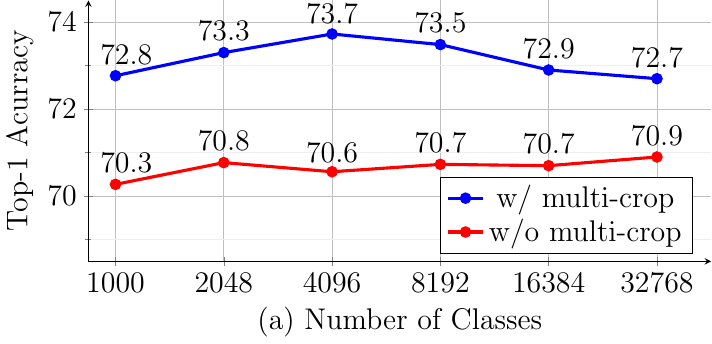}
        \includegraphics[width=0.9\linewidth]{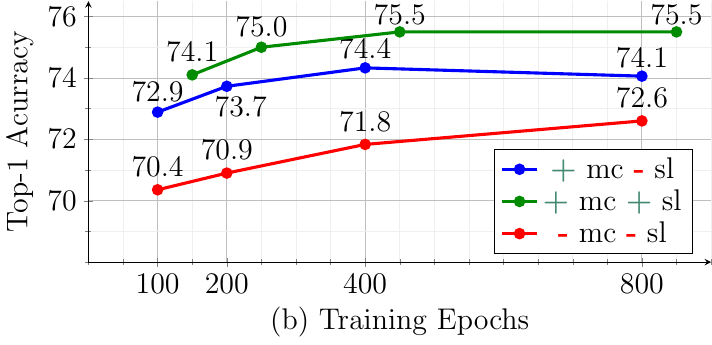}
        \end{center}
        \vspace{-0.4cm}
        \caption{\textbf{(a)} Effect of different numbers of classes in \ourmethod. \textbf{(b)} Effect of different training epochs in \ourmethod, where "mc" denotes multi-crop and "sl" denotes self-labeling. All results are ImageNet one-crop top-1 accuracy.}
        \vspace{-0.6cm}
\label{cls_epochs}
\end{figure}

\textbf{Number of Classes:} We show the impact of class number $C$ in Fig.~\ref{cls_epochs} (a). To make a comprehensive evaluation, we show the results of \ourmethod with and without multi-crop. The models are trained by setting the number of classes from 1000 to 32768. With multi-crop, \ourmethod performs best when the number of classes is 4096. Overall, the performances are quite stable and fluctuate within the range of 1\%, particularly when without multi-crop and $C >= 2048$. 

\textbf{Training Epochs:} Fig.~\ref{cls_epochs} (b) shows the performances of training \ourmethod with different epochs. Training longer improves the performance of \ourmethod model without multi-crop, while has less impact on the \ourmethod model with multi-crop (when the training epochs $>$ 400).

\section{Conclusion}

In this paper, we have presented a novel self-supervised approach \ourmethod. With a single loss function, our method can learn to classify images without labels, reaching 40.6\% top-1 accuracy on ImageNet. The learned representations can be used in a wide range of downstream tasks to achieve better results than existing state-of-the-art methods, including linear classification, semi-supervised classification, transfer learning, and dense prediction tasks such as object detection and instance segmentation. \ourmethod is simple and theoretically explainable. It does not rely on any clustering tools, making it easy to implement. There are many topics worth exploring in future work such as extensions to other modalities, and applications of \ourmethod to larger datasets.

\textbf{Broader Impact:} \ourmethod is a self-supervised method that tries to capture the intrinsic semantic structure from input datasets. Therefore, the learned model may be vulnerable to data distributions. With biased datasets, the model is likely to learn malicious information. The issue should be taken considered when using this method.

\textbf{Limitation:} When using ViTs as backbones, we adopt the momentum encoder to improve the performances, making the training strategy inconsistent with CNNs. In the future, we will explore strategies to remove the momentum encoder for ViTs without degrading performance.

{\small
\bibliographystyle{ieee_fullname}
\bibliography{twist}
}
\newpage
\appendix

\section{PyTorch-like Pseudo Code}
We give the PyTorch-like pseudo code, shown in Algorithm~\ref{alg:cap}.
\begin{algorithm}[h]
\caption{\ourmethod Pseudocode in a PyTorch-like style}
\label{alg:cap}
\definecolor{codeblue}{rgb}{0.25,0.5,0.5}
\definecolor{codekw}{rgb}{0.85, 0.18, 0.50}
\lstset{
  backgroundcolor=\color{white},
  basicstyle=\fontsize{7.5pt}{7.5pt}\ttfamily\selectfont,
  columns=fullflexible,
  breaklines=true,
  captionpos=b,
  commentstyle=\fontsize{7.5pt}{7.5pt}\color{codeblue},
  keywordstyle=\fontsize{7.5pt}{7.5pt}\color{codekw},
}
\vspace{-0.2cm}
\begin{lstlisting}[language=python]
# loss function
def twist_loss(p1, p2, alpha=1, beta=1):
    # calculate the consistency term: KL-divergence
    kl_div = ((p2*p2.log()).sum(dim=1) - 
                 (p2*p1.log()).sum(dim=1)).mean()
                 
    # calculate the sharpness term
    mean_ent = -(p1*p1.log()).sum(dim=1).mean()
    
    # calculate the diversity term
    mean_prob = p1.mean(dim=0)
    ent_mean = -(mean_prob * mean_prob.log()).sum()
    
    return kl_div + alpha*mean_ent - beta*ent_mean
    
# f: encoder network: backbone + projection head.
# bn: batch normalization operation with no affine parameters.
# B: batch-size.
# C: class number.
for x in loader:  # load a minibatch
    x1, x2 = aug(x), aug(x)  # random augmentation
    feat1, feat2 = f(x1), f(x2)

    p1 = softmax(bn(feat1), dim=1)
    p2 = softmax(bn(feat2), dim=1)
    
    # symmetric twist loss
    L = 0.5*(twist_loss(p1, p2) + twist_loss(p2, p1)
    L.backward()  # back-propagate
    update(f.param)  # SGD update
\end{lstlisting}
\end{algorithm}

\section{Visualization of feature similarity}
\begin{figure}[t]
   \begin{center}
   \includegraphics[width=1.0\linewidth]{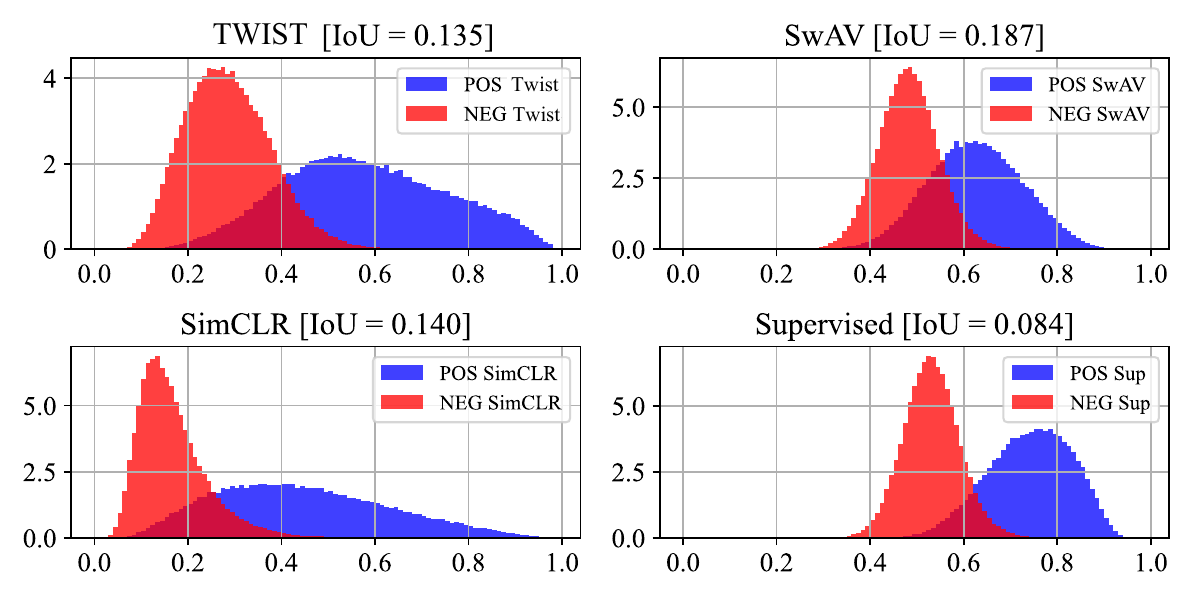}
   \end{center}
  \vspace{-0.4cm}
   \caption{The distributions of positive/negative similarities. IoU is the area of overlap between positive and negative distributions.}
   \label{relation}
\end{figure}

Fig. \ref{relation} shows the similarities of features sampled from the same or different classes of ImageNet. Specifically, we collect the outputs of the backbone as features, and calculate the cosine similarities. For positive samples, we sample two images from the same ImageNet class. For negative samples, we sample two images from different ImageNet classes. We then l2-normalize the features and calculate the similarities of positive/negative samples. The similarity distributions are shown in Fig. \ref{relation}. We compare \ourmethod with SimCLR~\cite{chen2020simple}, SwAV~\cite{caron2020unsupervised} and Supervised models~\cite{he2016deep}. From Fig. \ref{relation}, we observe that the postive distributions and the negative distributions of \ourmethod are more separable than other self-supervised methods. 

\section{Implementation Details}
\subsection{Self-supervised Pre-training}
The projection head is an non-linear MLP consisting of three layers with dimensions of $4096-4096-C$. The first two layers of the projection head are followed by a batch normalization and rectified linear units. After the projection head, we add a batch-normalization layer without affine parameters and finally a softmax operation to calculate the probability distributions. 
For multi-crop, we set the global scale to (0.4, 1.0) and local scale to (0.05, 0.4). For CNN backbones, we use 12 crops. For  ViTs, we use 6 crops for DeiT-S, and 10 crops for ViT-B~\cite{caron2021emerging}.

To train CNN backbones, we use LARS optimizer \cite{you2017scaling} with a cosine annealing learning rate schedule \cite{loshchilov2016sgdr}. We use a batch-size of 2048 splitting over 16 Tesla-V100 GPUs for ResNet50, and a batch-size of 1920 splitting over 32 Tesla-V100 GPUs for ResNet50$\times 2$. The learning rate is set to $0.5 \times$ $B$ $/256$. The weight decay is set to 1.5e-6. The computation and memory costs are mainly from the multi-crop augmentations. For model without multi-crop, 8 Tesla-V100 GPUs are enough to achieve 72.6\% top-1 linear accuracy on ImageNet. 
For self-labeling, we choose the samples in a mini-batch whose classification confidence (the maximum of softmax output) is larger than a predefined threshold. In practice, we use a cosine annealing schedule to choose the top 50\% to 60\% confident samples. We then use the chosen samples to generate hard labels for the subsequent fine-tuning process. For augmentation, we use multi-crop augmentations same as the regular training setting, except for that we change the global crop scale to (0.14, 0.4) and the local crop scale to (0.05, 0.14). We use the predictions of global crops as labels to guide the learning of local crops. 

For ViT-based models, we use the momentum encoder to enable stable training. The momentum is set to a variable value raised from 0.996 to 1 with a cosine annealing schedule. We change the weight of the three different terms of \ourmethod loss. Specifically, we set the coefficient as $\alpha=0.4$, $\beta=1.0$ for the sharpness term and diversity term, respectively. We evaluate the linear performance using the momentum encoder, similar as \cite{caron2021emerging} (with 0.1 performance improvement compared with the online network). 
With momentum encoder, the training objective is the assymetric form of:

\hspace{-0.2cm}
$\mathcal{L}(\bar{P}^{1}, P^{2}) =$
\vspace{-0.7cm}
\begin{equation}
    \small
    \begin{aligned}
    \hspace{-0.2cm}
         \\ 
        \underbrace{\frac{1}{B} \sum_{i=1}^{B} D_{KL}(\bar{P}^{1}_{i}||P^{2}_{i})}_{\text{consistency term}}  +  \underbrace{\frac{\alpha}{B}\sum_{i=1}^B H(P^{2}_{i})}_{\text{sharpness term}}  - \beta \underbrace{H(\frac{1}{B} \sum_{i=1}^{B} P_{i}^{2})}_{\text{diversity term}},
    \end{aligned}
    \label{form: kl}
\end{equation}
where $\bar{P}^1$ is the output of the momentum encoder, and $\bar{P}^2$ is the output of the online network.
The optimizer is AdamW \cite{loshchilov2017decoupled}. We use a batch-size of 1024 splitting over 16 Tesla-V100 GPUs. The learning rate is set to $0.0003 \times$ $B$ $/256$ for DeiT-S and $0.00075\times$ $B$ $/256$ for ViT-B. The weight decay is set to 0.06-0.12 with a cosine annealing strategy for Deit-S and 0.06 for ViT-B, we use no drop path for DeiT-S and drop path rate 0.1 for ViT-B.

\subsection{Unsupervised Classification} 
Our \ourmethod model can be regarded as a clustering function which takes images as input and directly outputs the assignments. Therefore, we adopt the measures for clustering in evaluation, including normalized mutual information (NMI), adjusted mutual information (AMI), and adjusted rand index (ARI). Besides, we map the predicted assignments to the class labels of ImageNet to evaluate the unsupervised classification accuracy. We use the Kuhn–Munkres algorithm \cite{kuhn1955hungarian} to find the best one-to-one permutation mapping, following the settings in IIC \cite{Ji_2019_ICCV} and SCAN \cite{van2020scan}. \textit{Though this process uses the real labels, they do not participate in any training process and are only used to map the prediction to the meaningful ImageNet labels.} We use ResNet-50 as backbone and set $C=1000$ in accordance with ImageNet.

\section{Examples for unsupervised classification}
\begin{figure*}
\centering
\centering
\includegraphics[width=0.8\linewidth]{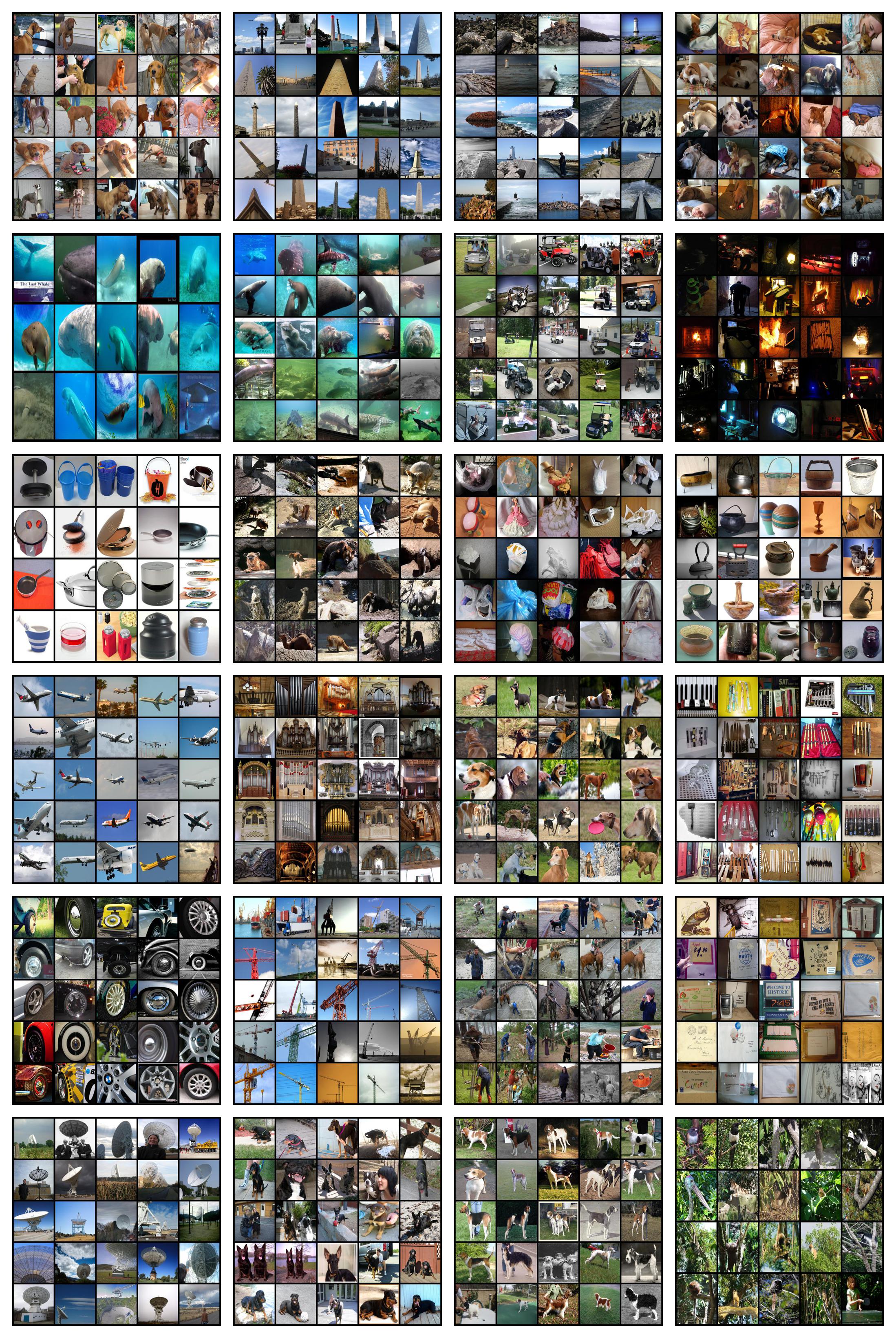}
\caption{Randomly chosen classes. We \textit{\textbf{randomly}} choose 24 classes for visualization. For each class, we \textit{\textbf{randomly}} choose 25 pictures to display. Note we did not make any selections on the pictures or the categories, all the categories and images are randomly chosen to give readers the accurate impression.}
\label{fig:randomcls}
\end{figure*}

\begin{figure*}
\hspace*{-0.14cm}\hspace*{0.05cm}\includegraphics[width=0.5\linewidth]{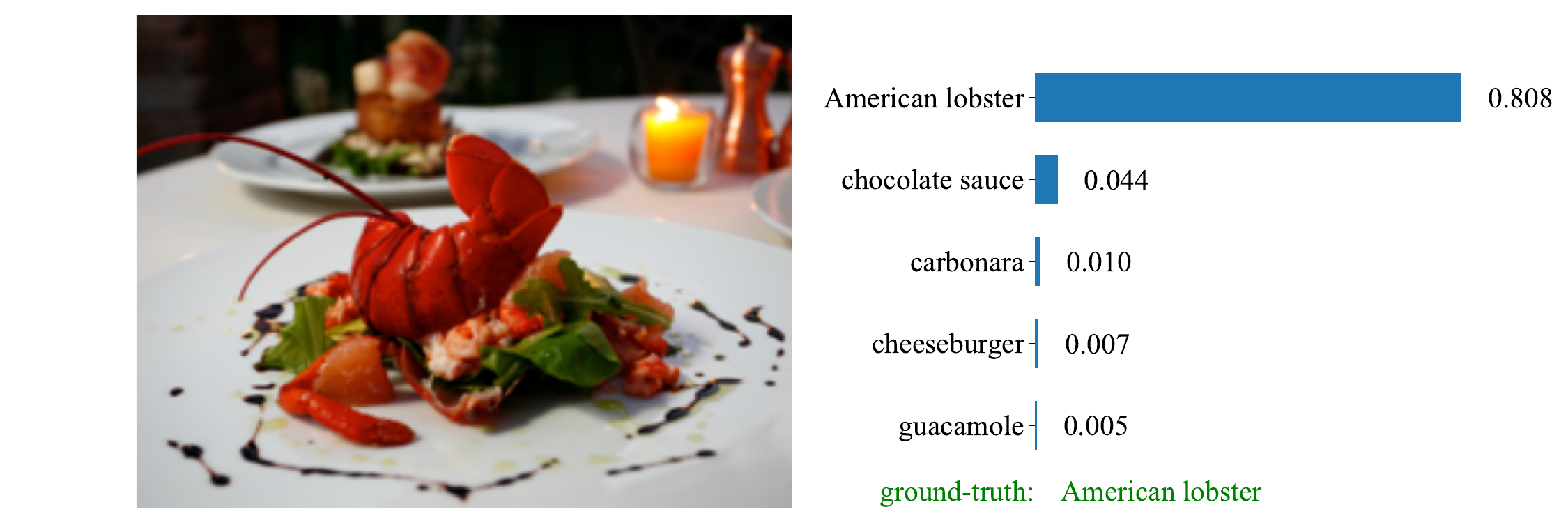}
\hspace*{-0.1cm}\hspace*{0.39cm}\includegraphics[width=0.5\linewidth]{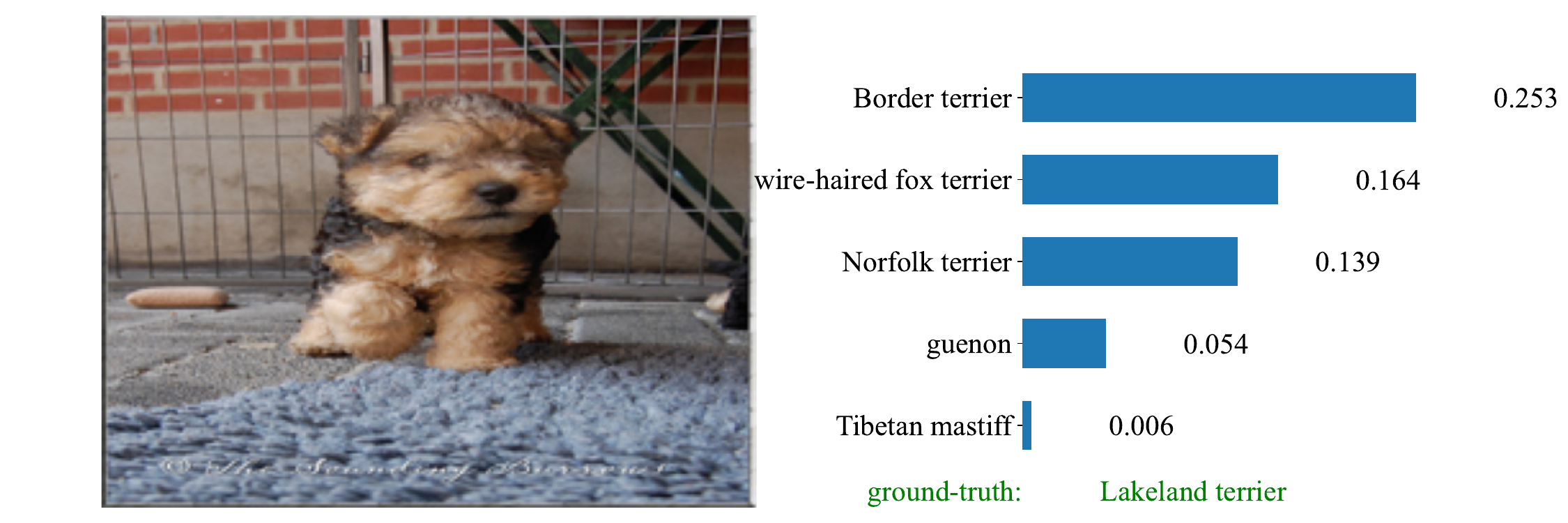}
\hspace*{-0.2cm}\includegraphics[width=0.5\linewidth]{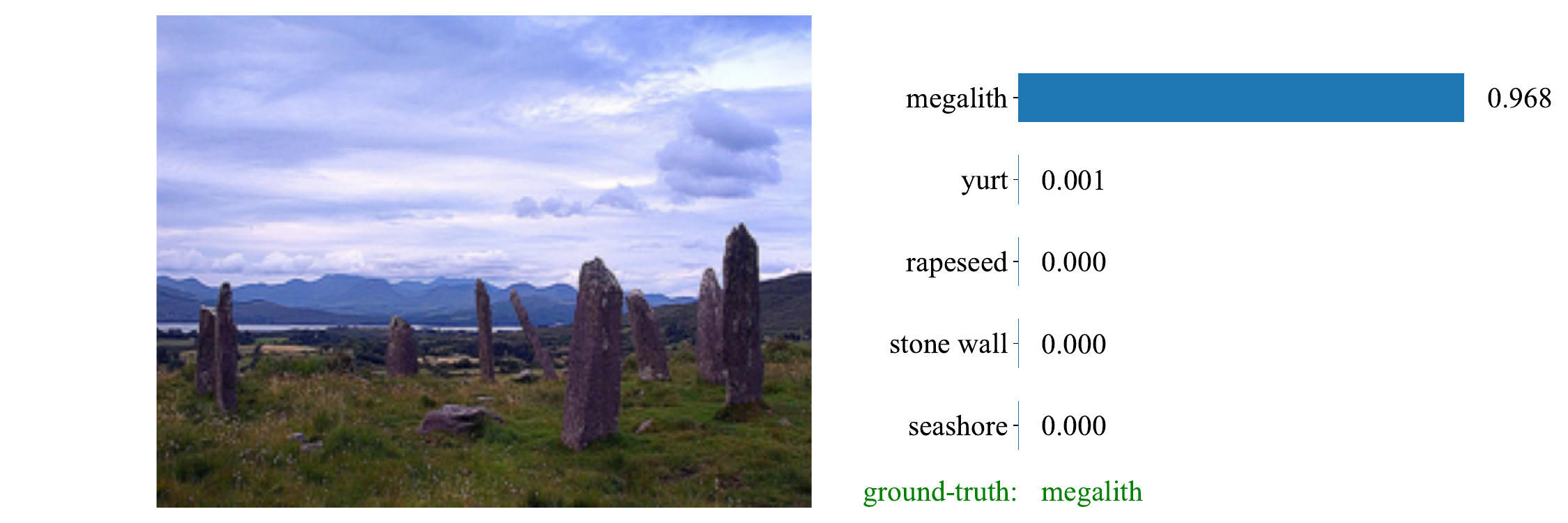}
\hspace*{-0.17cm}\hspace*{0.04cm}\includegraphics[width=0.5\linewidth]{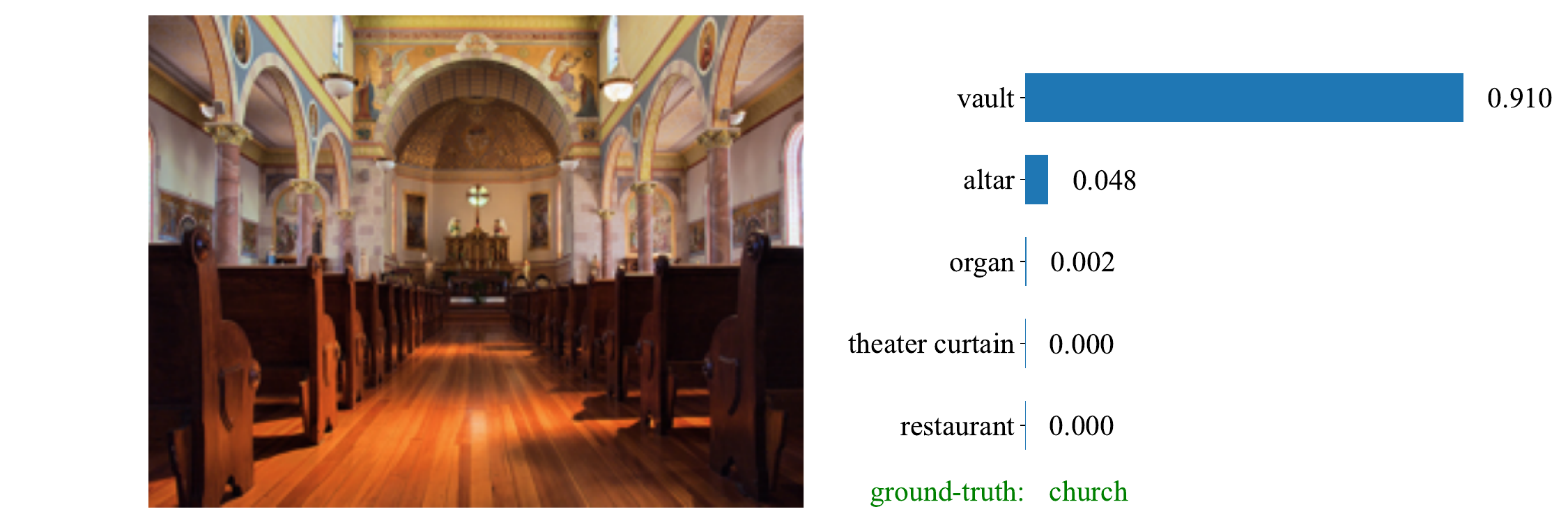}
\hspace*{-0.2cm}\includegraphics[width=0.5\linewidth]{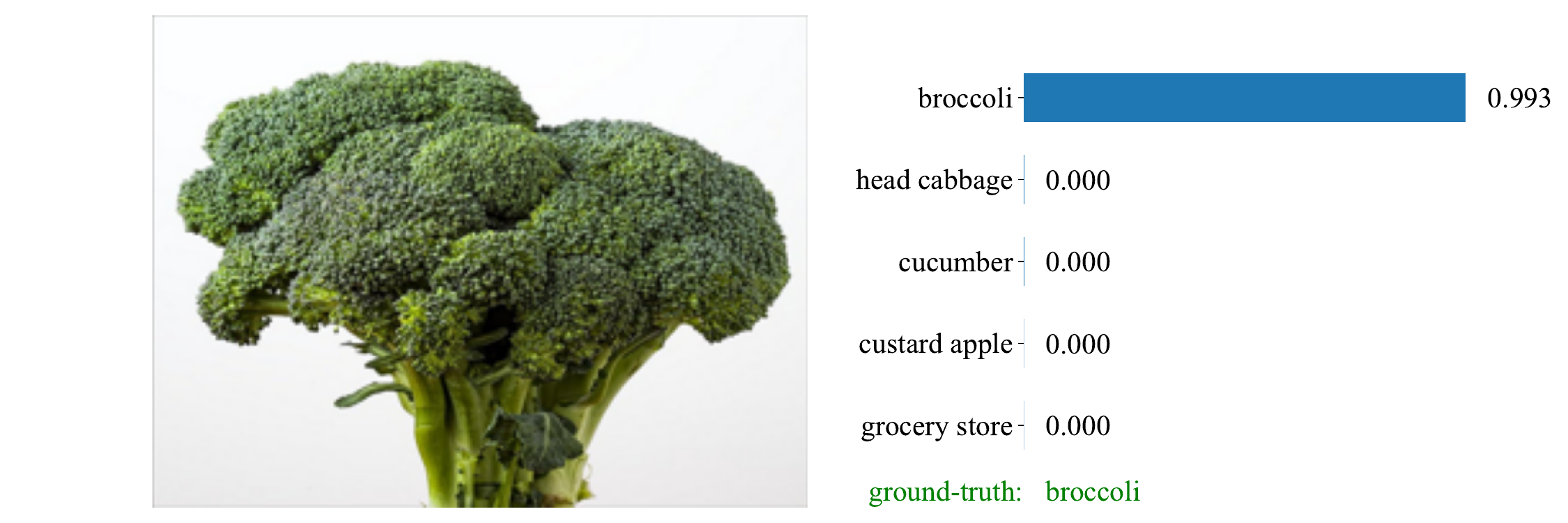}
\hspace*{-0.2cm}\hspace*{0.15cm}\includegraphics[width=0.5\linewidth]{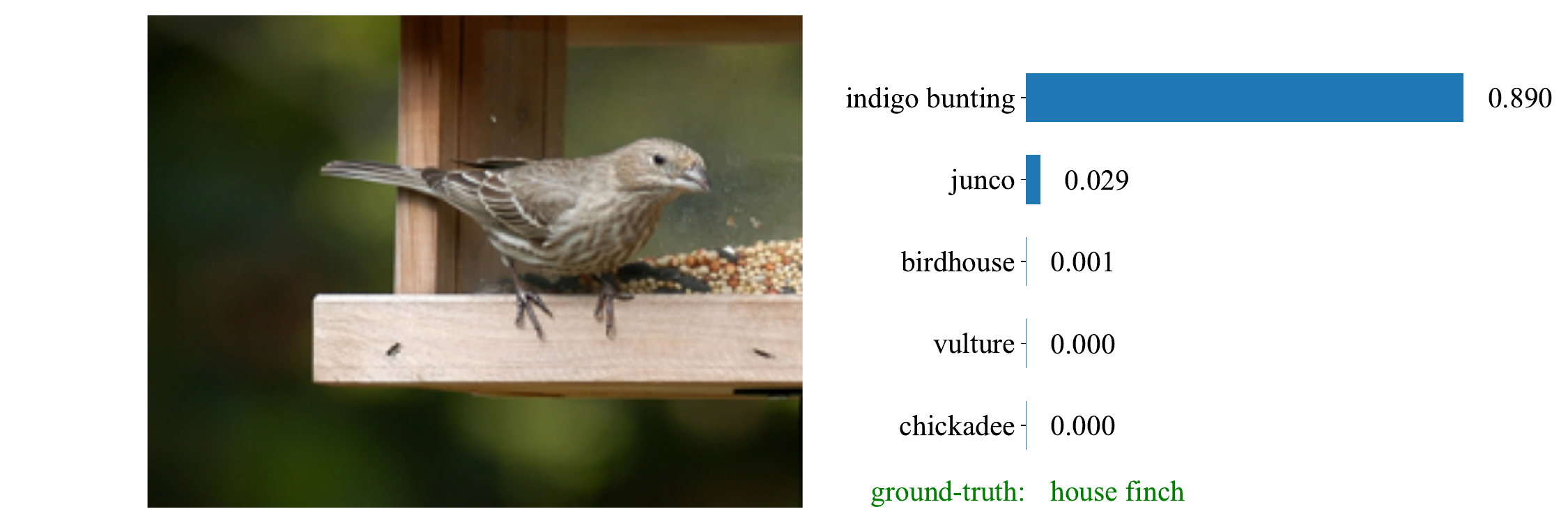}
\hspace*{0.0cm}\hspace*{-0.1cm}\includegraphics[width=0.5\linewidth]{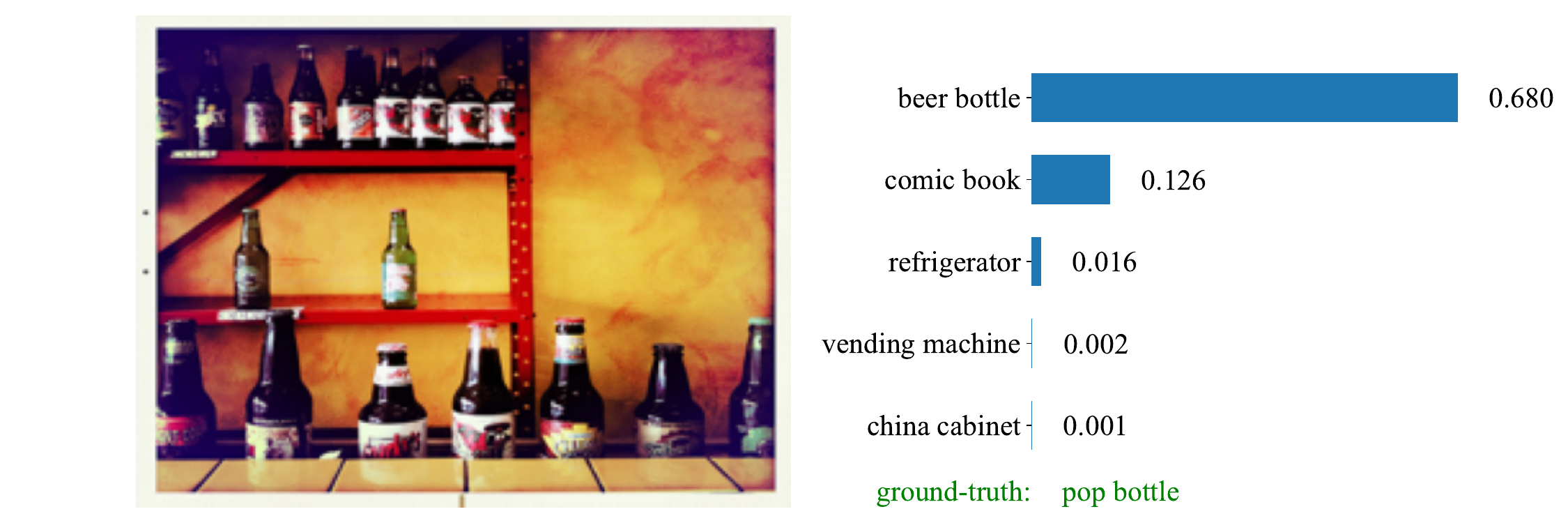}
\hspace*{0.11cm}\includegraphics[width=0.5\linewidth]{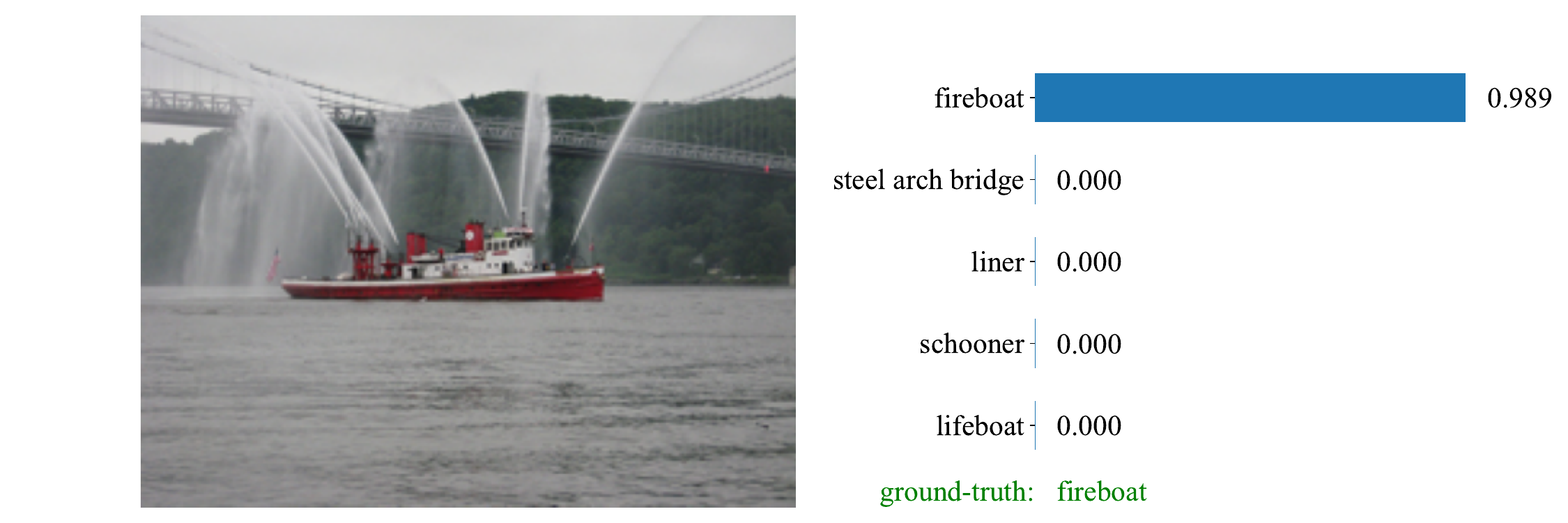}
\hspace*{0.0cm}\hspace*{-0.1cm}\includegraphics[width=0.5\linewidth]{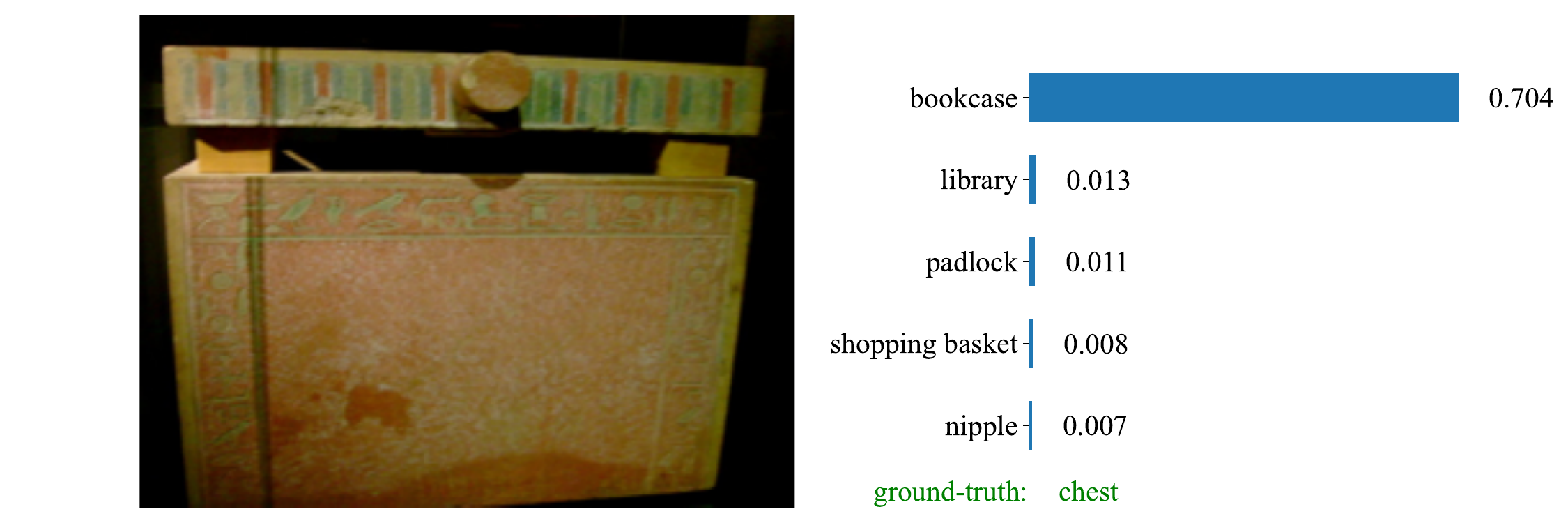}
\hspace*{-0.0cm}\hspace*{0.21cm}\includegraphics[width=0.5\linewidth]{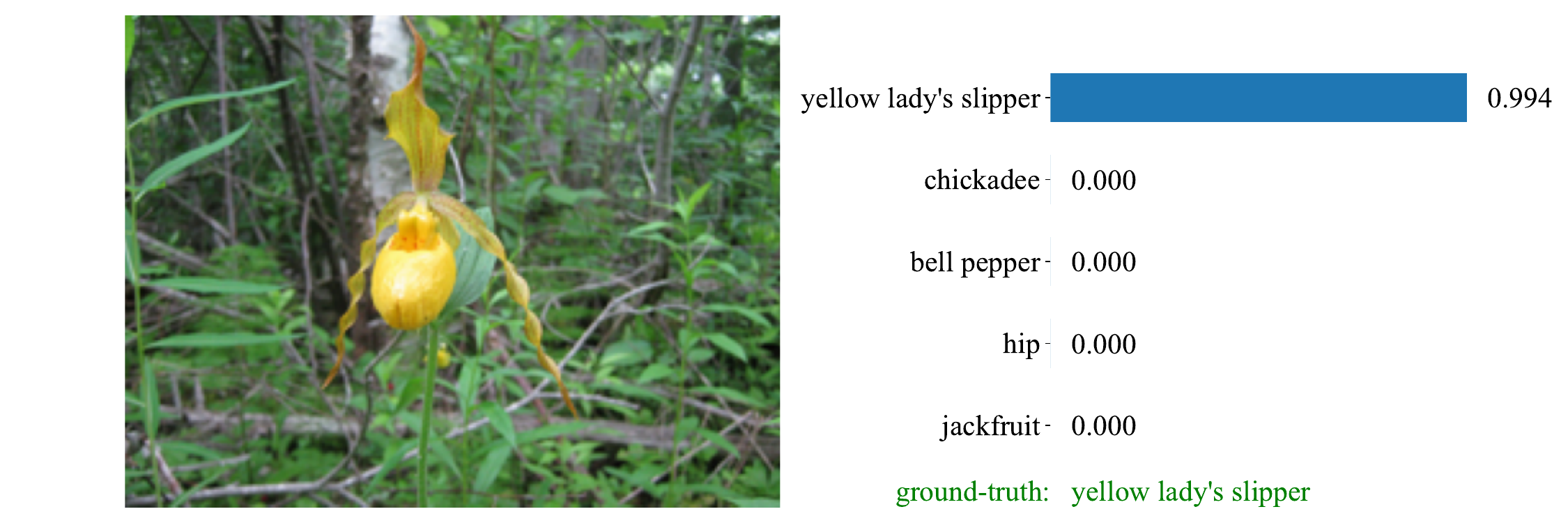}
\hspace*{0.12cm}\hspace*{-0.1cm}\includegraphics[width=0.5\linewidth]{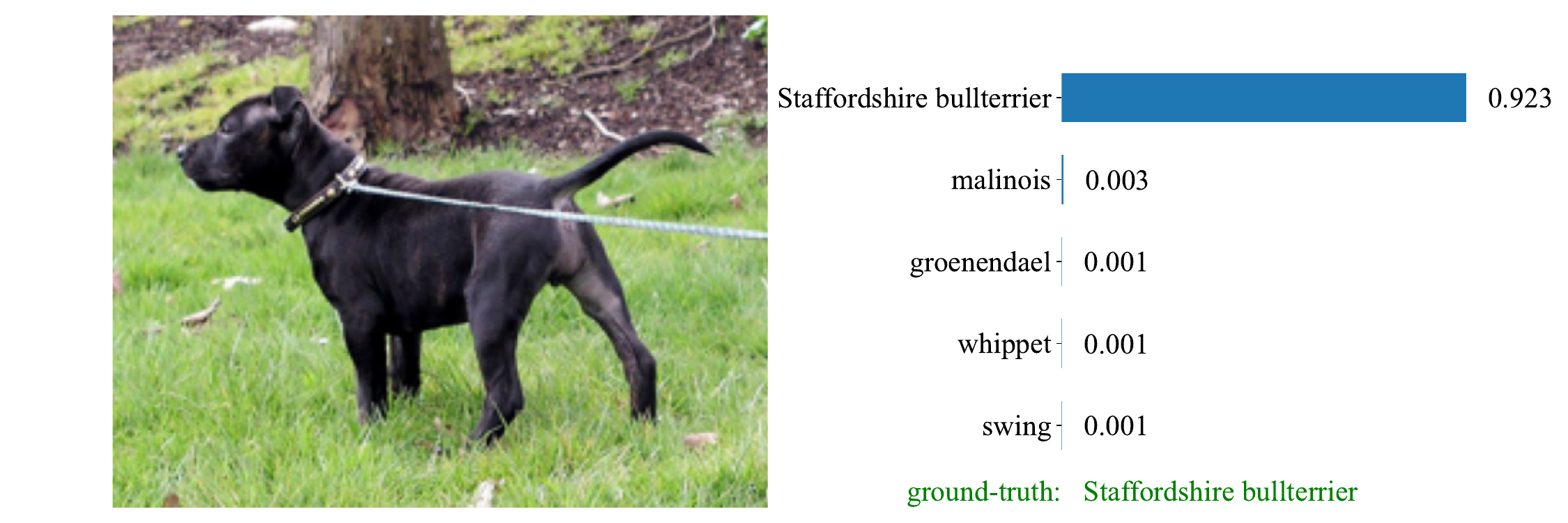}
\hspace*{-0.7cm}\includegraphics[width=0.5\linewidth]{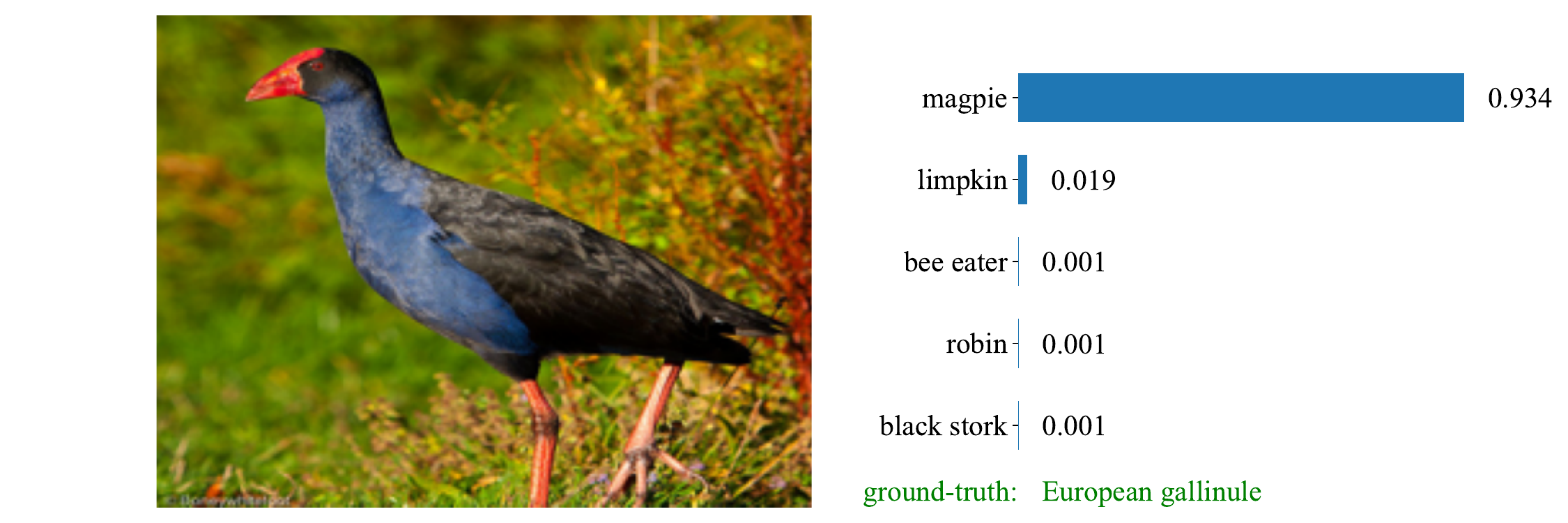}
\hspace*{-0.1cm}\hspace*{-0.1cm}\includegraphics[width=0.5\linewidth]{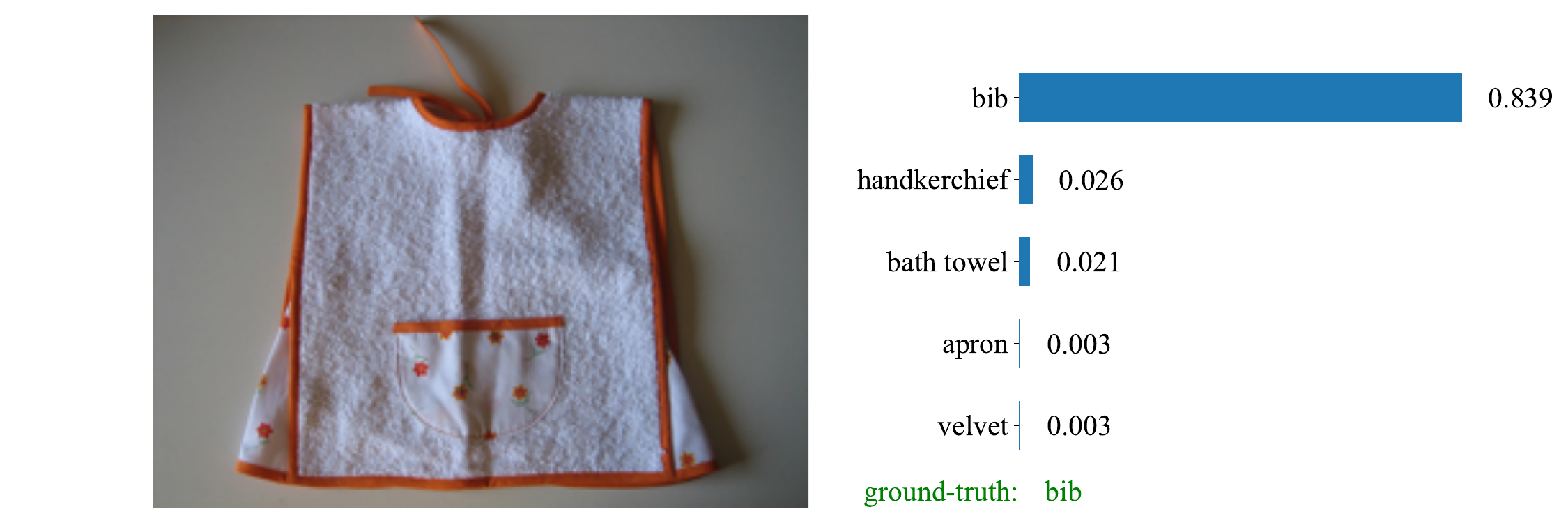}
\hspace*{0.3cm}\hspace*{-0.07cm}\includegraphics[width=0.5\linewidth]{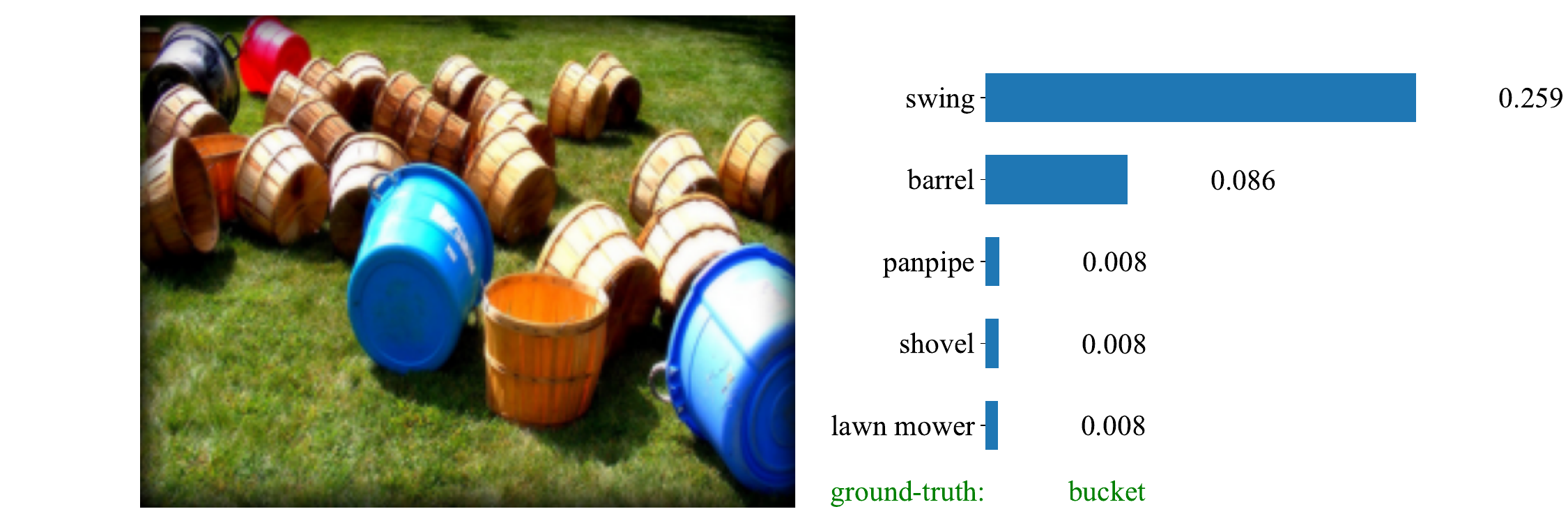}

\caption{Top-5 ImageNet accuracy of unsupervised classification. The labels are mapped by the Kuhn–Munkres algorithm.}
\label{fig:top5}
\end{figure*}

To give a qualitative impression on the performance of unsupervised classification, we display the learned partitions as Fig.~\ref{fig:randomcls}. We also display the top-5 predictions of some randomly selected pictures, shown as Fig.~\ref{fig:top5}. Specifically, the labels are mapped to the labels in ImageNet by Kuhn–Munkres algorithm. \textit{Note that the labels are only used to map our predictions to the meaningful ImageNet label descriptions, we do not use any label to participate in the training process.}

\section{Results and Analyses on Dense Tasks.}
Table \ref{det} shows the object detection performance on Pascal VOC and COCO datasets and the instance segmentation performance on COCO. We compare different architectures, namely C4 and FPN. From Table \ref{det}, we find some interesting phenomenons. \textbf{(1)} With architectures using feature pyramid network, \ourmethod achieves state-of-the-art results. Clustering-based methods also perform pretty well. \textbf{(2)} For C4 architectures, \ourmethod and clustering-based methods perform worse than contrastive learning methods like MoCo-v2. 

We thought the reason of the above phenomenons is that the classification-based methods (\ourmethod and clustering-based methods) tend to capture category-level invariances instead of instance-level invariances, which makes the outputs of the last convolutional layer discard intra-class variations that is useful for dense predictive tasks. When using FPN-based detectors, features of different layers are combined to compensate for the discarded information from the last layer. Less work concentrates on the effect of the intermediate layers of self-supervised models, while we find the intermediate features may preserve useful information for dense tasks. When using the FPN detectors, \ourmethod even outperforms those self-supervised methods designed specifically for the dense tasks, such as DenseCL \cite{wang2020DenseCL}, i.e., +1.6 AP on COCO detection. 

We find the similar phenomenon on semantic segmentation, shown in Table \ref{seg}. We give results of semantic segmentation with different architectures. The FCN architecture has no fusion of different layers, while FCN-FPN and DeepLab v3+ have the fusion operation. From Table \ref{seg}, we could observe that when combining with feature pyramid network, our method achieves best or competitive results. Without fusion of different layers of features, \ourmethod and other clustering-based models perform worse than contrastive learning methods.
\section{Change Log}
The results of ViTs are updated in Section 5. The linear result of DeiT-S/16 is improved from 75.6\% to 76.3\%, and the linear result of Vit-B is improved from 77.3\% to 78.4\%. The improved results are due to the more adequate hyper-parameter searching. Specifically for DeiT-S:
\begin{itemize}
    \item We change the batch-size from 2048 to 1024.
    \item The learning rate is changed from 0.0005 to 0.0003.
    \item The weight decay is changed from 0.06 to a cosine scheduler from 0.06 to 0.12.
\end{itemize}
For ViT-B: 
\begin{itemize}
    \item We change the batch-size from 2048 to 1024.
    \item We change drop path rate from 0.0 to 0.1.
\end{itemize}

\captionsetup{width=.85\textwidth}

\captionsetup{width=1.0\textwidth}

\captionsetup{width=.75\textwidth}
\begin{table*}[t]
\begin{center}
\begin{tabular}{l|ccc|ccc|ccc}
\Xhline{2\arrayrulewidth}
\multirow{2}{*}{Method}  & \multicolumn{3}{c|}{VOC07+12 det} & \multicolumn{3}{c|}{COCO det} & \multicolumn{3}{c}{COCO instance seg} \\
 & AP$_{\rm all}$ & AP$_{\rm 50}$ & AP$_{\rm 75}$ & AP$^{\rm bb}_{\rm all}$ & AP$^{\rm bb}_{\rm 50}$ & AP$^{\rm bb}_{\rm 75}$ & AP$^{\rm mk}_{\rm all}$ & AP$^{\rm mk}_{\rm 50}$ & AP$^{\rm mk}_{\rm 75}$ \\
\Xhline{2\arrayrulewidth}
\multicolumn{1}{l}{\textbf{\textit{C4}}} \\
\Xhline{2\arrayrulewidth}
Sup & 53.5 & 81.3 & 58.8 & 38.2 & 58.2 & 41.2 & 33.3 & 54.7 & 35.2 \\
Moco-v2 & \bf 57.4 & 82.5 & \bf 64.0 & \bf 39.3 & 58.9 & \bf 42.5 & 34.4 & 55.8 & 36.5 \\
SimCLR $^\dagger$& 57.0 & 82.4 & 63.5 & 38.5 & 58.5 & 41.7 & 33.8 & 55.1 & 36.0 \\
SwAV & 56.1 & \bf 82.6 & 62.7 & 38.4 & 58.6 & 41.3 & 33.8 & 55.2 & 35.9 \\
DINO $^\dagger$& 55.2 & 81.8 & 61.3 & 37.4 & 57.8 & 40.0 & 33.0 & 54.3 & 34.9 \\
DC-v2 $^\dagger$ & 54.2 & 81.6 & 59.9 & 37.0 & 57.7 & 39.5 & 32.8 & 54.2 & 34.4 \\
SimSiam & 57.0 & 82.4 & 63.7 & 39.2 & \bf 59.3 & 42.1 & 34.4 & 56.0 & 36.7 \\
BarlowTwins & 56.8 & \bf 82.6 & 63.4 & 39.2 & 59.0 & \bf 42.5 & 34.3 & 56.0 & 36.5 \\
\ourmethod & 55.3 & 82.2 & 61.2 & 38.0 & 58.4 & 40.8 & 33.5 & 54.9 & 35.5 \\
\Xhline{2\arrayrulewidth}
\multicolumn{1}{l}{\textbf{\textit{FPN}}}\\
\Xhline{2\arrayrulewidth}
Moco-v2 & 56.4 & 81.6 & 62.4 & 39.8 & 59.8 & 43.6 & 36.1 & 56.9 & 38.7 \\
SimCLR $^\dagger$& \bf 58.2 & 83.8 & 65.1 & 41.6 & 61.8 & 45.6 & 37.6 & 59.0 & 40.5 \\
SwAV & 57.2 & 83.5 & 64.5 & 41.6 & 62.3 & \bf 45.7 & \bf 37.9 & 59.3 & \bf 40.8 \\
DC-v2 $^\dagger$& 57.0 & 83.7 & 64.1 & 41.0 & 61.8 & 45.1 & 37.3 & 58.7 & 39.9 \\
DINO $^\dagger$& 57.2 & 83.5 & 63.7 & 41.4 & 62.2 & 45.3 & 37.5 & 58.8 & 40.2 \\
DenseCL & 56.9 & 82.0 & 63.0 & 40.3 & 59.9 & 44.3 & 36.4 & 57.0 & 39.2 \\
\rowcolor{backcolor} \ourmethod & 58.1 & \bf 84.2 & \bf 65.4 & \bf 41.9 & \bf 62.6 & \bf 45.7 & \bf 37.9 & \bf 59.7 & 40.6\\
\Xhline{2\arrayrulewidth}
\end{tabular}
\end{center}
\vspace{-0.4cm}
\caption{Object detection and instance segmentation results. For methods marked with $^\dagger$, we download the pre-trained models and run the detection and segmentation by ourselves. We report results both with C4 architecture and FPN architecture. For VOC dataset, we run 5 times and report the average.}
\label{det}
\vspace{0.4cm}
\begin{center}
\begin{tabular}{l|cc|cc|cc}
\Xhline{2\arrayrulewidth}
\multirow{2}{*}{Method}  & \multicolumn{2}{c|}{FCN} & \multicolumn{2}{c|}{FCN-FPN} & \multicolumn{2}{c}{DeepLab-v3} \\
 & VOC & Cityscapes & VOC & Cityscapes & VOC & Cityscapes  \\
\Xhline{2\arrayrulewidth}
Rand init & 40.7 & 63.5 & 37.9 & 62.9 & 49.5 & 68.3 \\ 
Sup & 67.7 & 73.7 & 70.4 & 75.4 & 76.6 & 78.6 \\
Moco-v2 & 67.5 & 74.5 & 67.5 & 75.4 & 72.9 & 78.6 \\
SimCLR $^\dagger$ & 68.9 & 72.9 & 72.8 & 74.9 & 78.5 & 77.8 \\
SwAV & 66.4 & 71.4 & 71.9 & 74.4 & 77.2 & 77.0 \\
DC-v2 $^\dagger$& 65.6 & 70.8 & 72.1 & 73.8 & 76.0 & 76.2 \\
DINO $^\dagger$& 66.0 & 71.1 & 71.9 & 73.8 & 76.4 & 76.2 \\
\rowcolor{backcolor} \ourmethod & 66.7 & 71.5 & 73.3 & 74.6 & 77.3 & 76.9\\
\Xhline{2\arrayrulewidth}
\end{tabular}
\end{center}
\vspace{-0.4cm}
\caption{Semantic segmentation with different architectures. All results are averaged over 5 trials.}
\label{seg}
\end{table*}

\end{document}